\documentclass{article}

\usepackage{microtype}
\usepackage{graphicx}
\usepackage{subfigure}
\usepackage{booktabs} % for professional tables

\usepackage{hyperref}

\usepackage[accepted]{icml2023}

\usepackage{amsmath}
\usepackage{amssymb}
\usepackage{mathtools}
\usepackage{amsthm}

\usepackage[capitalize,noabbrev]{cleveref}

\theoremstyle{plain}
\newtheorem{theorem}{Theorem}[section]
\newtheorem{proposition}[theorem]{Proposition}
\newtheorem{lemma}[theorem]{Lemma}
\newtheorem{corollary}[theorem]{Corollary}
\newtheorem{example}{Example}
\theoremstyle{definition}
\newtheorem{definition}[theorem]{Definition}
\newtheorem{assumption}[theorem]{Assumption}
\theoremstyle{remark}

\usepackage[textsize=tiny]{todonotes}

\usepackage{multirow}

\usepackage{tikz}
\usetikzlibrary{bayesnet}
\usetikzlibrary{arrows.meta}
\usepackage{subfigure}
\usepackage{wrapfig}

\tikzstyle{causallatent} = [circle, fill=white, draw=black, inner sep=1pt,
minimum size=18pt, font=\fontsize{8}{8}\selectfont, node distance=1]
\tikzstyle{causalobs} = [causallatent, fill=gray!25]

\tikzstyle{causaleps} = [rectangle, fill=white, draw=black, inner sep=1pt,
minimum size=18pt, font=\fontsize{8}{8}\selectfont, node distance=1]

\tikzstyle{causalobsset} = [ellipse, fill=white, draw=black, inner sep=1pt,
minimum size=18pt, font=\fontsize{8}{8}\selectfont, node distance=1, fill=gray!25]

\renewcommand{\edge}[3][]{ %
  \foreach \x in {#2} { %
    \foreach \y in {#3} { %
      \path (\x) edge [-To, #1] (\y) ;%
    } ;
  } ;
}

\newcommand{\uedge}[3][]{ %
  \foreach \x in {#2} { %
    \foreach \y in {#3} { %
      \path (\x) edge [-, red, #1] (\y) ;%

    } ;
  } ;
}

\usepackage{algorithm}
\usepackage{algorithmic}
\renewcommand{\algorithmicrequire}{\textbf{Input:}}
\renewcommand{\algorithmicensure}{\textbf{Output:}}
\usepackage{enumitem}
\usepackage{multicol}

\newcommand\ci{\perp\!\!\!\perp}

\usepackage{nicematrix}
\tikzset{highlight/.style={rectangle,
fill=red!15,
rounded corners = 1.0 mm,
inner sep=1pt,
fit=#1}}

\tikzset{highlight2/.style={rectangle,
fill=orange!15,
rounded corners = 1.0 mm,
inner sep=1pt,
fit=#1}}

\tikzset{solid-rectangle/.style={rectangle,
fill=none,
rounded corners = 1.0 mm,
draw=black,
inner sep=1.5pt,
fit=#1}}

\tikzset{dashed-rectangle/.style={rectangle,dashed,
fill=none,
rounded corners = 1.0 mm,
draw=black,
inner sep=1.5pt,
fit=#1}}

\begin{document}

\twocolumn[
\icmltitle{Causal Discovery with Latent Confounders Based on Higher-Order Cumulants}



\icmlsetsymbol{equal}{*}

\begin{icmlauthorlist}
\icmlauthor{Ruichu Cai}{gdut,pengcheng}
\icmlauthor{Zhiyi Huang}{gdut}
\icmlauthor{Wei Chen}{gdut}
\icmlauthor{Zhifeng Hao}{gdut,stu}
\icmlauthor{Kun Zhang}{cmu,mbzuai}
\end{icmlauthorlist}
\icmlaffiliation{gdut}{School of Computer Science, Guangdong University of Technology, Guangzhou, China}
\icmlaffiliation{pengcheng}{Peng Cheng Laboratory, Shenzhen, China}
\icmlaffiliation{stu}{College of Engineering, Shantou University, Shantou, China}
\icmlaffiliation{cmu}{Department of Philosophy, Carnegie Mellon University, Pittsburgh, PA, United States}
\icmlaffiliation{mbzuai}{Mohamed bin Zayed University of Artificial Intelligence, Abu Dhabi, United Arab Emirates}

\icmlcorrespondingauthor{Wei Chen}{chenweiDelight@gmail.com}

\icmlkeywords{Machine Learning, ICML}

\vskip 0.3in
]



\printAffiliationsAndNotice{}  % leave blank if no need to mention equal contribution

\begin{abstract}
Causal discovery with latent confounders is an important but challenging task in many scientific areas. Despite the success of some overcomplete independent component analysis (OICA) based methods in certain domains, they are computationally expensive and can easily get stuck into local optima. We notice that interestingly, by making use of higher-order cumulants, there exists a closed-form solution to OICA in specific cases, e.g., when the mixing procedure follows the One-Latent-Component structure. In light of the power of the closed-form solution to OICA corresponding to the One-Latent-Component structure, we formulate a way to estimate the mixing matrix using the higher-order cumulants, and further propose the testable One-Latent-Component condition to identify the latent variables and determine causal orders. By iteratively removing the share identified latent components, we successfully extend the results on the One-Latent-Component structure to the Multi-Latent-Component structure and finally provide a practical and asymptotically correct algorithm to learn the causal structure with latent variables. Experimental results illustrate the asymptotic correctness and effectiveness of the proposed method.
\end{abstract}

\section{Introduction}

Causal discovery with latent confounders is needed in many scientific discovery fields because we are usually unable to collect or measure all the underlying causal variables. Such latent confounders generally pose serious identifiability problems in learning causal structures \citep{spirtes2000causation, hyvarinen2013pairwise, chen2021causal, adams2021identification}. In this paper, we consider the problem of causal discovery with latent confounders in which observed variables may be directly dependent.

\begin{figure}[t]
\label{fig:intro_graph}
    \centering
    \subfigure[]{
        \begin{tikzpicture}[thick]
            \node[causallatent] (L) {$L$}; %
            \node[causalobs, xshift= -0.60cm, yshift=-1.20cm] (X1) {$X_1$} ;
            \node[causalobs, xshift=  0.60cm, yshift=-1.20cm] (X2) {$X_2$} ;
            \edge{L}{X1, X2};
            \draw (-0.625cm, -0.55cm) node [inner sep=0.75pt]    {$\alpha_{1}$};
            \draw ( 0.625cm, -0.55cm) node [inner sep=0.75pt]    {$\alpha_{2}$};
        \end{tikzpicture}
    }
    \hspace{0.5cm}
    \subfigure[]{
        \begin{tikzpicture}[thick]
            \node[causallatent, xshift= -0.60cm, yshift=  0.0cm] (L1) {$L_1$}; %
            \node[causallatent, xshift=  0.60cm, yshift=  0.0cm] (L2) {$L_2$}; %
            \node[causalobs,    xshift= -1.20cm, yshift= -1.20cm] (X1) {$X_1$} ; %
            \node[causalobs,    xshift=  0.00cm, yshift= -1.20cm] (X2) {$X_2$} ; %
            \node[causalobs,    xshift=  1.20cm, yshift= -1.20cm] (X3) {$X_3$} ; %

            \draw (-1.20cm, -0.55cm) node [inner sep=0.75pt]    {$\alpha_{1}$};
            \draw (-0.60cm, -0.55cm) node [inner sep=0.75pt]    {$\alpha_{2}$};
            \draw ( 0.00cm, -0.55cm) node [inner sep=0.75pt]    {$\alpha_{3}$};

            \draw ( 0.60cm, -0.55cm) node [inner sep=0.75pt]    {$\beta_{2}$};
            \draw ( 1.20cm, -0.55cm) node [inner sep=0.75pt]    {$\beta_{3}$};

            \edge {L1} {X1, X2, X3} ; %
            \edge {L2} {X2, X3} ; %
        \end{tikzpicture}
    }

    \caption{Two Directed Acyclic Graphs (DAGs), where the white node represents the latent variable and the grey node represents the observed variable.}
\end{figure}
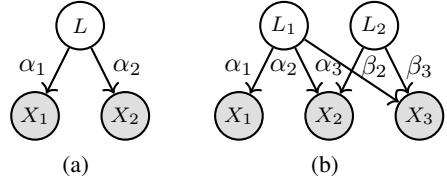

Various methods, such as FCI \citep{spirtes1995causal}, RFCI \citep{colombo2012learning}, and their variants \citep{colombo2014order} are proposed to solve the above problem, but they cannot go beyond the well-known Markov equivalence class. By utilizing linear and non-Gaussian models \citep{hoyer2008estimation, hyvarinen2013pairwise, tashiro2014parcelingam, chen2021causal, salehkaleybar2020learning, wang2020high, wang2020causal}, one is able to identify the causal structure even in the presence of latent confounders. One typical model is the Latent-Variable Linear, non-Gaussian Acyclic Model (lvLiNGAM) \citep{hoyer2008estimation}, which utilizes the Overcomplete Independent Component Analysis (OICA) \citep{eriksson2004identifiability, lewicki2000learning} to estimate the mixing matrix from the data, and then obtain the causal structure through the estimated mixing matrix. In practice, many existing OICA algorithms aim to maximize the likelihood of the data. As a result, OICA algorithms often rely on the Expectation-Maximization (EM) procedure along with approximate inference techniques, like Gibbs sampling \citep{olshausen1999learning} and mean-field approximation \citep{hojen2002mean}. Overall, OICA is hard to compute and may get stuck in local optima. 

Accordingly, one practical issue with causal discovery with latent confounders in the linear, non-Gaussian case is how to estimate the mixing matrix and then recover causal relationships without making use of the traditional OICA procedure. Interestingly, by assuming the causal model is linear and leveraging the higher-order cumulants that can be considered as a measure of certain types of non-Gaussianity \citep{comon2010handbook}, we provide a closed-form solution to OICA in specific cases. One specific case corresponds to the structure with two observed variables and one shared latent component directly affecting them, defined as the \textit{One-Latent-Component structure} and illustrated in Figure \ref{fig:intro_graph}(a). We find that by employing higher-order cumulants, the ratio of causal coefficients from the latent confounders can be estimated by  $\frac{\alpha_1}{\alpha_2}=\frac{cum(X_1, X_1, X_1, X_2)}{cum(X_1, X_1, X_2, X_2)}$, where $cum(X_1, X_1, X_1, X_2)$ denotes the cumulant of variables $(X_1, X_1, X_1, X_2)$. Moreover, if the variables have unit variance, we can estimate the causal coefficient from latent confounder $L$ to every observed variable with $cum(X_1, X_2)= \alpha_1 \alpha_2$ (note that $cum(X_1, X_2)$ is actually the covariance between $X_1$ and $X_2$). This observation can also be extended to more than one latent component case. 

In light of the power of the above procedure, we provide a method for estimating the mixing matrix when the causal structure is given. By exploring the relation among the exogenous and endogenous variables in the mixing matrix, we further develop a latent variable causal model estimation algorithm under mild assumptions. The main idea is as follows. First, we assume that the causal structure over two observed variable sets $\mathbf{X_i}$ and $\mathbf{X_j}$ is a One-Latent-Component structure. Then, we estimate the mixing matrix based on the hypothetical structure. Whether the hypothetical structure is true or not is examined by the proposed \textit{One-Latent-Component condition}. This condition holds for $\mathbf{X_i}$ and $\mathbf{X_j}$ if and only if there exists a non-zero weight vector $\omega$ such that $\omega^{\top}\mathbf{A}_{\mathbf{X_j},L}=\mathbf{0}$ and $\omega^{\top}\mathbf{X_j} \ci \mathbf{X_i}$, where $\mathbf{A}_{\mathbf{X_j}, L}$ is the sub-mixing matrix of $\mathbf{X_j}$ with a latent component $L$. This idea can be generalized to identify the Multi-Latent-Component structure by removing the shared independent components and further recover the full causal structure. 

The organization of this paper is as follows. Section \ref{sec:cumulant} provides a closed-form solution to OICA in specific cases. Section \ref{sec:model_estimation} introduces the One-Latent-Component condition and proposes a causal discovery method for the linear latent variable model. Section \ref{sec:experiments} provides the experiment results on synthetic and real-world data. Section \ref{sec:disscution} discusses the limitation of the proposed method in real-world applications. Section \ref{sec:conclusion} is the conclusion.

\section{Closed-Form Solution to OICA in Specific Cases}\label{sec:cumulant}

Linear OICA assumes the following data generating model:
\begin{equation}
    \mathbf{X} = \mathbf{A}\mathbf{S},
    \label{eq:ica}
\end{equation}
where $\mathbf{X} \in \mathbb{R}^{p}$, $\mathbf{S} \in \mathbb{R}^{d}$, and  $\mathbf{A} \in \mathbb{R}^{p \times d}$ are observed variables (mixtures), independent components, and mixing matrix respectively. In OICA, $d > p$. The goal of OICA is to recover $\mathbf{A}$ from $\mathbf{X}$. 

To develop a closed-form solution to OICA in specific cases, we begin with a simple structure where two variables share only one independent component, which is defined as the \textit{One-Latent-Component structure}, illustrated in Figure \ref{fig:intro_graph}(a).
\begin{definition}[One-Latent-Component structure] Let $\mathbf{X_i}$ and $\mathbf{X_j}$ be two observed variable sets following Eq. (\ref{eq:ica}). If $\mathbf{X_i}$ and $\mathbf{X_j}$ share only one latent component $L$, $\mathbf{X_i} \leftarrow L \rightarrow \mathbf{X_j}$ is a One-Latent-Component structure.
\end{definition}

Without loss of generality, in this paper we assume that all variables are zero-mean and that all latent variables have a unit variance. In a One-Latent-Component structure, the mixing matrix can be obtained based on the higher-order cumulants.
Before providing the closed-form solution, we introduce the definition of cumulant:
\begin{definition}
[Cumulant \citep{brillinger2001time}] Let $X=(X_1,X_2, \dots, X_n)$ be a random vector of length $n$. The $k$-th order cumulant tensor of $X$ is defined as a $n \times \cdots  \times n$ ($k$ times) table, $\mathcal{C}^{(k)}$, whose entry at position ($i_1, \cdots, i_k$) is

\begin{small}
\begin{equation}
    \begin{aligned}
    &\mathcal{C}^{(k)}_{i_1 \!, \cdots \!,  i_k}  = cum(X_{i_1}, \dots, X_{i_k}) \\
    &=  \sum_{(D_1\!, \dots \!, D_h)} (-1)^{h \!- \!1}(h \!- \!1)! \mathbb{E}\!\left[ \! \prod_{j\in D_i}X_j  \! \right] \cdots \mathbb{E} \!\left[ \! \prod_{j \in D_h} \! X_j \! \right]\!,
    \end{aligned}
\end{equation}
\end{small}

where the sum is taken over all partitions $(D_1, \dots, D_h)$ of the set $\{i_1, \dots, i_k\}$. 
\end{definition}

If each variable $X_i$ has zero mean, then the sum of the partitions with size 1 is 0 and can be omitted. For example, in Figure \ref{fig:intro_graph}(a), the 2nd and 4th order cumulants are:
\begin{equation}
    \begin{aligned}
    cum(X_1, X_2) = & \mathbb{E}[X_1X_2] = \alpha_1\alpha_2 \mathbb{E}[L^2], \textrm{~and}\\
    cum(X_1, X_1, X_2, X_2) =& \mathbb{E}[X_1^2X_2^2] - \mathbb{E}[X_1X_1]\mathbb{E}[X_2X_2] \\
    -& 2\mathbb{E}[X_1X_2]\mathbb{E}[X_1X_2]\\
    =& \alpha_1^2\alpha_2^2 (\mathbb{E}[L^4] - 3\mathbb{E}[L^2]).
    \end{aligned}
    \label{eq:4ordercum}
\end{equation}
Note that for some distributions, the $k$-th order cumulants of the variables will be zero. For example, when $X_i$ follows a Gaussian distribution, the third and higher-order cumulants of $X_i$ are zero.
Interesting, it is worth noting that all other distributions than the Gaussian one have an infinite number of non-zero cumulants \citep{feller1991introduction}. Hence, given any non-Gaussian distributions, we can always find non-zero higher-order cumulants. In other words, under the non-Gaussian assumption, if some cumulants are zeros, we can resort to the other higher-order cumulants. Without loss of generality, we only show the results based on the 2nd and 4th-order cumulants, but our methodology can be modified to work on other cumulants. Specifically, based on Eq. (\ref{eq:4ordercum}), if $S$ is non-Gaussian, the mixing coefficients can be identified by the 2nd and 4th order cumulants, which is summarized as Theorem \ref{th:shared-one-exogenous}. The detailed procedure for using other typical higher-order cumulants is provided in Appendix \ref{app:est_by_general_cum}.

\begin{theorem}
Let $X_i$ and $X_j$ be two observed variables following Eq. (\ref{eq:ica}). Suppose $X_i$ and $X_j$ follow the One-Latent-Component structure and $S$ is the only one shared non-Gaussian latent component and has a unit variance. Then the mixing coefficients between $\{X_i, X_j\}$ and $S$, denoted by $\hat{\alpha}_i$ and $\hat{\alpha}_j$ respectively, can be identified by the fourth-order cumulant as follows:
\begin{equation}
    \begin{aligned}
        \hat{\alpha}_i &= \sqrt{\frac{cum(X_i, X_i, X_j, X_j)}{cum(X_i, X_j, X_j, X_j)} \cdot cum(X_i, X_j)}, \\
                     % &= \sqrt{\frac{cum(X_i, X_i, X_i, X_j)}{cum(X_i, X_i, X_j, X_j)} \cdot cum(X_i, X_j)},\\
        \hat{\alpha}_j &= \frac{cum(X_i, X_j)}{\hat{\alpha}_i}. 
    \end{aligned}
    \label{eq:shared-one-exogenous}
\end{equation}
\label{th:shared-one-exogenous}
\end{theorem}

Theorem \ref{th:shared-one-exogenous} provides the closed-form solution for estimating the mixing matrix in the One-Latent-Component structure case. If there are two shared independent components illustrated in Figure \ref{fig:intro_graph} (b), can we still identify the mixing matrix? Interestingly, we find that when considering only $X_1$ and $X_2$, the causal coefficient of $L_1$ on $X_1$ and $X_2$ (denoted by the coefficients $\alpha_1$ and $\alpha_2$) can be estimated. $\alpha_3$ can be estimated in the same way by considering $X_1$ and $X_3$. But $X_2$ and $X_3$ are influenced by two independent components, as shown in the $\tikz[baseline=1.5ex] {\node[draw=black, minimum width=2.5ex, minimum height=2.5ex, anchor=center, shape=rectangle, rounded corners = 1.0 mm] at (2ex,2ex) {}; }$ part in Eq. (\ref{eq:cum4mixing}), which violates the condition in Theorem \ref{th:shared-one-exogenous}. Can we still identify $\beta_2$ and $\beta_3$?  

\begin{equation}
\begin{aligned}
\begin{bmatrix}
X_{1}\\
X_{2}\\
X_{3}\\
\end{bmatrix} 
&= 
\begin{bNiceMatrix}[create-medium-nodes]
\CodeBefore [create-cell-nodes]
\begin{tikzpicture} [name suffix = -medium]
\node [solid-rectangle = (2-1) (3-2)] {} ;
\end{tikzpicture}
\Body
\alpha_{1} & 0         & \gamma_{1} & 0 & 0\\
\alpha_{2} & \beta_{2} & 0 & \zeta_{2} & 0\\
\alpha_{3} & \beta_{3} & 0 & 0 & \eta_{3}\\
\end{bNiceMatrix}
\begin{bmatrix}
L_{1}\\
L_{2}\\
S_{X_{1}}\\
S_{X_{2}}\\
S_{X_{3}}\\
\end{bmatrix}\\
\stackrel{
\omega
}{\Rightarrow}
\begin{bmatrix}
X_1 \\
\widetilde{X}_{2}\\
X_{3}\\
\end{bmatrix}
&=
\begin{bNiceMatrix}[create-medium-nodes]
\CodeBefore [create-cell-nodes]
\begin{tikzpicture} [name suffix = -medium]
\node [dashed-rectangle = (2-2) (3-2)] {} ;
\end{tikzpicture}
\Body
\alpha_{1} & 0         & \gamma_{1} & 0 & 0\\
0          & \beta_{2} & -\frac{\alpha_2}{\alpha_1}\gamma_{1} & \zeta_{2} & 0\\
\alpha_{3} & \beta_{3} & 0 & 0 & \eta_{3}\\
\end{bNiceMatrix}
\begin{bmatrix}
L_{1}\\
L_{2}\\
S_{X_{1}}\\
S_{X_{2}}\\
S_{X_{3}}\\
\end{bmatrix}
\end{aligned}
\label{eq:cum4mixing}
\end{equation}

One may think if we can remove the influence of $L_1$ from $X_2$, then $\widetilde{X}_2$ (the residual after removing the influence of $L_1$ on $X_2$) and $X_3$ share only one component $L_2$. Specifically, let $\omega$ be a non-zero weight vector satisfying $\omega^{\top}[\alpha_1, \alpha_2]^\top=0$, then $\widetilde{X}_2 = \omega^\top [X_1, X_2]^\top$ and $X_3$ share one latent component $L_2$, as seen from the $\tikz[baseline=1.5ex] {\node[draw=black, minimum width=2.5ex, minimum height=2.5ex, anchor=center, shape=rectangle, dashed, rounded corners = 1.0 mm] at (2ex,2ex) {}; }$ part in Eq. (\ref{eq:cum4mixing}). So the mixing coefficients $\beta_2$ and $\beta_3$ can be further estimated directly by Eq. (\ref{eq:shared-one-exogenous}). Intuitively, we have $\widetilde{X}_2 \ci L_1$, because $\widetilde{X}_2$ successfully removes the influence of $L_1$ by making use of $X_1$ as a surrogate, which will be explained below. This particular regression is formalized as follows.

\begin{definition}
[Surrogate Regression] Let $X_j$ be a single observed variable and $\mathbf{X_k}$ be an observed variable set following Eq. (\ref{eq:ica}). Suppose $\mathbf{S'}$ be the shared independent components for $X_j$ and $\mathbf{X_k}$, and $\mathbf{X_k}$ be the surrogate variable set of $\mathbf{S'}$. By using $\mathbf{X_k}$, the influence from $\mathbf{S'}$ to $X_j$ can be removed. Then the \emph{surrogate regression residual} of $X_j$ on $\mathbf{S'}$ by utilizing $\mathbf{X_k}$ as a set of surrogate variables is defined as:
\begin{equation}
\begin{aligned}
    \widetilde{X}_j &= \omega^{\top}[X_j, \mathbf{X_k}]^\top\\
    &= 
    \omega^{\top}
    \begin{bmatrix}
        \mathbf{A}_{{\{X_j,\mathbf{X_k}\}, \mathbf{S'}}} & \mathbf{A}_{{\{X_j,\mathbf{X_k}\}, \mathbf{S''}}}\\
    \end{bmatrix}
    \begin{bmatrix}
        \mathbf{S'}\\
        \mathbf{S''}\\
    \end{bmatrix}\\
    &= 
    \omega^{\top} \mathbf{A}_{\{X_j,\mathbf{X_k}\}, S''} \mathbf{S''},
\end{aligned}
    \label{eq:mixingres}
\end{equation}
where $\mathbf{A}_{{\{X_j,\mathbf{X_k}\}, \mathbf{S'}}}$ is the sub-matrix of $\mathbf{A}$ in Eq. (\ref{eq:ica}), which represents the mixing coefficient from $\mathbf{S'}$ to $\{X_j,\mathbf{X_k}\}$, and $\omega$ satisfies $\omega^{\top}\mathbf{A}_{\{X_j,\mathbf{X_k}\},\mathbf{S'}} = \mathbf{0}$ and $\omega \neq \mathbf{0}$.
\label{def:remove}
\end{definition}

Note that in this specific case, we can estimate the mixing matrix $\mathbf{A}$ from data by Eq. (\ref{eq:shared-one-exogenous}). Based on this definition and motivated by the above example, we can recursively recover the mixing matrix according to Theorem \ref{th:shared-one-exogenous}, which is summarized as Theorem \ref{th:shared-multiple-exogenous}.

\begin{theorem}
Let $X_i$ and $X_j$ be two observed variables and $\mathbf{X_k}$ be an observed variable set following Eq. (\ref{eq:ica}) (where $X_i, X_j \notin \mathbf{X_k}$), $\mathbf{S_1}$ be an independent component set and $S_2$ be an independent component not in $\mathbf{S_1}$. Suppose $\{\mathbf{S_{1}},  S_{2} \}$ are the shared independent components of $X_i$ and $X_j$, and that $\mathbf{S_{1}}$ are the shared independent components of $\mathbf{X_k}$ and $\lbrace X_i, X_j \rbrace$. Then the mixing coefficient from $S_2$ to $\{X_i, X_j\}$ can be identified when $X_i$ and $\widetilde{X}_j$ share only one independent component $S_2$, where $\widetilde{X}_j$ is the \emph{surrogate regression residual} of $X_j$ on $\mathbf{S_1}$ by utilizing $\mathbf{X_k}$ as surrogates.
\label{th:shared-multiple-exogenous}
\end{theorem}

\section{Estimation of Canonical lvLiNGAM}\label{sec:model_estimation}

In this section, we aim to extend the above results in specific cases to estimate the canonical lvLiNGAM with some mild constraints by an efficient strategy. Specially, we will begin with the canonical lvLiNGAM and provide the connection between lvLiNGAM and OICA, and then intuitively show how to use the results of specific OICA to identify the latent variables and determine the causal order in a canonical lvLiNGAM, and finally provide the estimation method for the canonical lvLiNGAM. 

\subsection{Canonical lvLiNGAM}
A canonical lvLiNGAM is $\mathbf{X} =\mathbf{BX} + \mathbf{\Lambda} \mathbf{L} +\mathbf{S_X}$, where $\mathbf{X}$ is the vector of the observed variables, $\mathbf{L}$ is the vector of the latent confounders, $\mathbf{S_X}$ represents the {\it independent non-Gaussian noises}, and $\mathbf{B}$ and $\mathbf{\Lambda}$ are the causal coefficient matrices of $\mathbf{X}$ and $\mathbf{L}$ respectively. We also assume $\mathbf{\Lambda}$ has full column rank. According to the definition of canonical lvLiNGAM \cite{hoyer2008estimation}, both the latent variables and noises are exogenous variables, which can be regarded as mutually independent components in OICA. Relating the canonical lvLiNGAM to the OICA, the model can be rewritten as:
\begin{equation}
    \mathbf{X}  =\begin{bmatrix}
                (\mathbf{I} - \mathbf{B})^{-1} \mathbf{\Lambda}  & (\mathbf{I} -\mathbf{B})^{-1}
                \end{bmatrix}\begin{bmatrix}
                \mathbf{L} \\
                \mathbf{S_X}
                \end{bmatrix}
                = \mathbf{A}\mathbf{S},
    \label{eq:mixing_matrix}
\end{equation}
where $\mathbf{A} = \begin{bmatrix}(\mathbf{I} - \mathbf{B})^{-1} \mathbf{\Lambda} & (\mathbf{I} -\mathbf{B})^{-1} \end{bmatrix}$ is the mixing matrix and $\mathbf{S} = [\mathbf{L}, \mathbf{S_X}]^{\top}$ is the independent components. Both $\mathbf{L}$ and $\mathbf{S_X}$ are special cases of $\mathbf{S}$.

To estimate the canonical lvLiNGAM, we further make the following assumptions. 

\begin{assumption}\label{asm:pure}
[At least one pure set of observed variables] There is at least one pure set of observed variables for a latent confounder as a child. A pure set of observed variables is a nonempty subset of observed variables that have no direct connections to other variables, except for at most one latent confounder. But we allow causal edges among the observed variables within the same pure set of observed variables.
\end{assumption}

\begin{assumption}\label{asm:threechild}
[Three-child variables] Each latent confounder has at least three observed variables as children.
\end{assumption}

\begin{assumption}\label{asm:faithful}
[Causal faithfulness \citep{spirtes2000causation}] Suppose $X_i$ and $X_j$ be two observed variables and $\mathbf{C}$ is a subset of observed variables. Let $\mathbf{C} \bigcap \{X_{i}, X_{j} \} = \emptyset$. If $X_{i}$ and $X_{j}$ are independent conditional on $\mathbf{C}$ in $P$, then $X_{i}$ is d-separated from $X_{j}$ conditional on $\mathbf{C}$ in causal graph $\mathcal{G}$.
\end{assumption}

The practicality and rationality of the assumptions are as follows. Assumption \ref{asm:pure} is weaker than the pure measurement variable assumption in Tetrad \citep{silva2006learning}, Triad \citep{cai2019triad} and GIN \citep{xie2020generalized}, as they require each observed variable to be pure while our assumption of at least one pure set of observed variables is much weaker; the former implies the latter, and we allow causal edges among the variables in the same pure set.  Assumption \ref{asm:threechild} is also implied in the Triad and GIN conditions.  Assumption \ref{asm:faithful} is widely used in the existing causal discovery methods. 

\subsection{One-Latent-Component Condition}

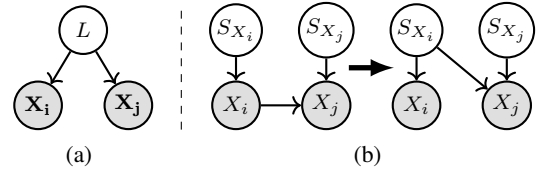
\begin{figure}
    \centering
    \subfigure[]{
        \begin{tikzpicture}[thick]
            \node[causallatent] (L) {$L$}; %
            \node[causalobsset, xshift= -0.6cm, yshift=-1cm] (X1) {$\mathbf{X_i}$} ;
            \node[causalobsset, xshift=  0.6cm, yshift=-1cm] (X2) {$\mathbf{X_j}$} ;
            \edge{L}{X1, X2};
            \end{tikzpicture}
    }
    \hspace{0.1cm}
    \begin{tikzpicture}
        \path (0.0cm, 0.4cm) edge [dashed] (0.0cm, 2.0cm);
    \end{tikzpicture}
    \hspace{0.1cm}
    \subfigure[]{
        \begin{tikzpicture}[thick]
            \node[causallatent, xshift= -1.8cm] (SX1-1) {$S_{X_i}$}; %
            \node[causallatent, xshift= -0.6cm] (SX2-1) {$S_{X_j}$}; %
            \node[causalobsset, xshift= -1.8cm, yshift=-1cm] (X1-1) {$X_i$} ;
            \node[causalobsset, xshift= -0.6cm, yshift=-1cm] (X2-1) {$X_j$} ;
            \edge{SX1-1}{X1-1};
            \edge{SX2-1}{X2-1};
            \edge{X1-1}{X2-1};
            \path (-0.3cm, -0.5cm) edge [-latex, line width=2px] (0.3cm, -0.5cm);
            \node[causallatent, xshift=  0.6cm] (SX1) {$S_{X_i}$}; %
            \node[causallatent, xshift=  1.8cm] (SX2) {$S_{X_j}$}; %
            \node[causalobsset, xshift=  0.6cm, yshift=-1cm] (X1) {$X_i$} ;
            \node[causalobsset, xshift=  1.8cm, yshift=-1cm] (X2) {$X_j$} ;
            \edge{SX1} {X1, X2};
            \edge{SX2} {X2};
        \end{tikzpicture}
    }
    \caption{Two Typical One-Latent-Component Structures. (a) A structure over two observed variable sets $\mathbf{X_i}$ and $\mathbf{X_j}$ which are influenced by 1 latent confounder $L$; (b) A structure over $X_i$ and $X_j$ which are influenced by 1 latent noise $S_{X_i}$ of $X_i$, transformed from the ground truth $X_i\to X_j$.}
    \label{fig:p1fs}
\end{figure}

To utilize the closed-form solution to OICA in specific cases to recover the causal model, including the structure, we consider a simple case with two variable sets $\mathbf{X_i}$ and $\mathbf{X_j}$ sharing only one independent component, i.e., the case with the One-Latent-Component structure.

According to the definition of the One-Latent-Component structure, we find that if the influence of the independent component $L$ on two variables $X_1$ and $X_2$ is removed, then these two variables are independent. Take Figure \ref{fig:intro_graph}(a) as an example. The influence of $L$ cannot be removed when only $X_2$ is utilized. If we utilize $X_1$ and $X_2$ to remove the influence of $L$, we can not certify from the data that we remove the influence of $L$ correctly. So we need to use an additional observed variable to give a testable condition, which can be used to verify that the structure between the observed variables is consistent with the hypothetical structure. Take Figure \ref{fig:intro_graph}(b) as an example. Let $\mathbf{X_i}=\{X_1\}$ and $\mathbf{X_j}=\{X_2, X_3\}$. We assume that $\mathbf{X_i}$ and $\mathbf{X_j}$ follow the \textit{One-Latent-Component structure}, where $\mathbf{X_i}$ and $\mathbf{X_j}$ share only one latent component $L_1$. Then we estimate the mixing matrix by the method provided in Section \ref{sec:cumulant}. Based on Eq. (\ref{eq:cum4mixing}) and Eq. (\ref{eq:mixingres}), we can find that $\omega^{\top}\mathbf{X_j}$ is independent of $\mathbf{X_i}$ because $\omega^{\top}\mathbf{X_j}$ is proportional to $[\alpha_3, -\alpha_2] [X_2, X_3]^{\top} = (\alpha_3 \beta_2 - \alpha_2 \beta_3 )L_2 + \alpha_3 \zeta_2 S_{X_2} - \alpha_2 \eta_{3} S_{X_3}$. Motivated by this, we introduce the definition of the \textit{One-Latent-Component condition} and then provide Theorem \ref{th:condition} to test whether $(\mathbf{X_i}, \mathbf{X_j})$ satisfies the condition.

\begin{definition}[One-Latent-Component condition] Let $\mathbf{X_i}$ and $\mathbf{X_j}$ be two subsets of observed variables following the canonical lvLiNGAM model. Suppose the dimension of $\mathbf{X_j}$ is greater than 2. $(\mathbf{X_i}, \mathbf{X_j})$ satisfies the One-Latent-Component condition if and only if there exists a non-zero $\omega$ such that $\omega^{\top}\mathbf{\widehat{A}}_{\mathbf{X_j},L}=\mathbf{0}$ and $\omega^{\top}\mathbf{X_j} \ci \mathbf{X_i}$, where $\mathbf{\widehat{A}}_{\mathbf{X_j}, L}$ is the causal coefficients of $L$ on $\mathbf{X_j}$ that can be estimated according to Eq. (\ref{eq:shared-one-exogenous}).
\label{def:1-component}
\end{definition}

\begin{theorem}
Suppose Assumptions \ref{asm:pure} - \ref{asm:faithful} hold. Let $\mathbf{X_i}$ and $\mathbf{X_j}$ be two dependent subsets of observed variables following the canonical lvLiNGAM model. $(\mathbf{X_i}, \mathbf{X_j})$ satisfies One-Latent-Component condition if and only if $\mathbf{X_i}$ and $\mathbf{X_j}$ follow the One-Latent-Component structure.
\label{th:condition}
\end{theorem}

Theorem \ref{th:condition} gives the conditions through a mixing matrix consistent with the hypothetical structure. In Figure \ref{fig:p1fs}, the mixing matrices for the models in (b) have the same support as that for the model in (a), but we can successfully distinguish between them. The methods for identifying the two structures are shown in Section \ref{sec:latent_variable} and Section \ref{sec:causal_order}.

\subsubsection{Identify the latent confounders}\label{sec:latent_variable}

To identify the latent confounders, we provide the following corollary based on the One-Latent-Component condition:

\begin{corollary} [Latent Confounder Identification] Suppose Assumptions \ref{asm:pure} - \ref{asm:faithful} hold. Let $\mathbf{X_i}$ and $\mathbf{X_j}$ be disjoint and dependent subsets of the observed variables following the canonical lvLiNGAM model. If $(\mathbf{X_i}, \mathbf{X_j})$ satisfies the One-Latent-Component condition, then $\mathbf{X_i}$ and $\mathbf{X_j}$ are directly caused by one latent confounder and there is no directed path between $\mathbf{X_i}$ and $\mathbf{X_j}$.
\label{pro:non-descendant}
\end{corollary}

Based on the above corollary, we can utilize the following steps to identify the latent confounders between observed variable sets $\mathbf{X_i}$ and $\mathbf{X_j}$ as follows: 1) estimate the mixing matrix $\widehat{\mathbf{A}}$, under the hypothetical structure that $\mathbf{X_i} \leftarrow L \rightarrow \mathbf{X_j}$ given in Figure \ref{fig:p1fs}(a), 2) if the estimated $\widehat{\mathbf{A}}$ satisfies the One-Latent-Component condition, then accept the hypothetical structure $\mathbf{X_i} \leftarrow L \rightarrow \mathbf{X_j}$ as well as the estimated mixing matrix $\widehat{\mathbf{A}}$, 3) otherwise, reject the hypothesis. 

\subsubsection{Determine the causal order}\label{sec:causal_order}

After identifying the latent confounders, the following question is: if two observed variables $X_i$ and $X_j$ are causally connected (i.e., one is a cause or ancestor of the other), how can we determine the causal order between them? A general method is trying to regress one on the another and test whether the residual is independent of the regressor. But if those two variables are affected by a latent confounder and directly connected, the asymmetry of this independent information is lost. However, one natural solution is if we can remove the influence of latent confounders, then they will share only one latent noise. That is, $X_i \to X_j$ can be transformed to the One-Latent-Component structure $X_i \gets S_{X_i} \to X_j$, as shown in Figure \ref{fig:p1fs}(b). We can derive the following corollary:

\begin{corollary} [Causal Order Determination] Suppose Assumptions \ref{asm:pure} - \ref{asm:faithful} hold. Let $X_i$ and $X_j$ be two dependent observed variables following the canonical lvLiNGAM model. If $(\{ X_i \}, \{X_i, X_j\})$ satisfies the One-Latent-Component condition, then $X_i$ is causally earlier than $X_j$.
\label{pro:descendant}
\end{corollary}

Based on the above corollary, we can utilize the following steps to determine the causal order between observed variable $X_i$ and $X_j$:  1) estimate the mixing matrix $\widehat{\mathbf{A}}$, under the hypothetical structure that $X_i \leftarrow S_{X_i} \rightarrow X_j, (i.e., X_i \rightarrow X_j)$ given in Figure \ref{fig:p1fs}(b); 2) if the estimated $\widehat{\mathbf{A}}$ satisfies the One-Latent-Component condition, then accept the hypothetical structure $X_i \rightarrow X_j$ as well the estimated mixing matrix $\widehat{\mathbf{A}}$, 3) otherwise, reject the hypothesis. 

\subsection{From One-Latent-Component Structure to Multi-Latent-Component Structure}\label{sec:alg}

In this section, we consider more general structures than the One-Latent-Component structure and provide an algorithm to estimate the canonical lvLiNGAM model. The framework of the algorithm is summarized as Algorithm \ref{alg:algorithm_framework}. First, we begin with an undirected complete causal graph over observed variables. Second, we identify the latent confounders by utilizing the One-Latent-Component condition according to Corollary \ref{pro:non-descendant}, i.e., if a One-Latent-Component structure is accepted, the causal graph and the corresponding mixing matrix are updated accordingly. The details are shown in lines 4-9 of Algorithm \ref{alg:algorithm_framework}. Third, similar to the above steps, we orient the causal direction of the edges by utilizing the One-Latent-Component condition according to Corollary \ref{pro:descendant}, as shown in lines 11-16 of Algorithm \ref{alg:algorithm_framework}. Finally, we eliminate redundant edges as shown in line 18 of Algorithm \ref{alg:algorithm_framework} (i.e., cutting off the edges between observed variables that are not adjacent). Due to space limitations, a running example is provided in Appendix \ref{app:toy}.

This algorithm is mainly based on the iterative identification of One-Latent-Component structures, which ensures the following two issues: 1) how to handle the cases that are not the One-Latent-Component structure, i.e., with the Multi-Latent-Component structure, and 2) how to eliminate the redundant edges (figuring out whether observed variables are adjacent)? Detailed solutions to these two key issues are introduced in the following two subsections.

\subsubsection{Remove shared identified latent components}

In this subsection, we will provide a way to remove the influence of the shared independent components with the help of surrogate variables, following the idea of Theorem \ref{th:shared-multiple-exogenous}. This may allow discovering the cases of more than one latent component structure with the help of the One-Latent-Component condition. The subsequent problem is how to choose surrogate variables from observed variables. 

Interestingly, with the help of the discovered One-Latent-Component structure, we notice that the two variables sets $\mathbf{X_i}$ and $\mathbf{X_j}$ in One-Latent-Component structure can help remove the information of the shared one latent component $L$. So we can use $\mathbf{X_i}$ and $\mathbf{X_j}$ as surrogate variables of each other to remove the influence of $L$. 

Therefore, considering two observed variable set $\mathbf{X_i}$ and $\mathbf{X_j}$, and their found shared components $\mathbf{S'}$, if we would like to detect whether there is another One-Latent-Component structure over them, then we can find a set of observed variables $\mathbf{X}_k$ that are also influenced by $\mathbf{S'}$, then $\mathbf{X}_k$ can be used as surrogate variables to ``regress" $\mathbf{X_j}$ according to Eq. (\ref{eq:mixingres}) to obtain $\omega^{\top} \mathbf{X_j}$. This can remove the influence of $\mathbf{S'}$, which helps to reveal previously unidentified shared latent components. The correctness of this procedure is guaranteed by the following theorem, which ensures that all shared independent components are identifiable under the model assumptions.

\begin{lemma}
Suppose Assumptions \ref{asm:pure} - \ref{asm:faithful} hold. If there exist unidentified shared independent components, there must exist a pair of observed variable sets $\mathbf{X_i}$ and $\mathbf{X_j}$, and the surrogate variables of their identified independent components $\mathbf{S'}$ such that $(\mathbf{X_i}, \mathbf{\widetilde{X}_j})$ satisfies the One-Latent-Component condition, where $\mathbf{\widetilde{X}_j}$ are the surrogate regression residuals of $\mathbf{X_j}$ on $\mathbf{S'}$.
    \label{th:ext-one-latent-component}
\end{lemma}

\begin{algorithm}[tb]
  \caption{Estimating the canonical lvLiNGAM model}
  \label{alg:algorithm_framework}
\begin{algorithmic}[1]
  \REQUIRE Observed variable set $\mathbf{X}$
  \ENSURE Causal graph $\mathcal{G}$ and mixing matrix $\widehat{\mathbf{A}}$
  \STATE Initialize $\mathcal{G}$ as a complete undirected graph, the mixing matrix $\widehat{\mathbf{A}} = \mathbf{0}$, and $\mathbf{V}$ as the variables that are non-fully directed-edge connected;
  \REPEAT
    \STATE //Identify the Latent Confounders
    \FOR{each pair of variables, $\{X_i, X_j\} \in \mathbf{V}$}
        \STATE $\widetilde{X}_i, \widetilde{X}_j \leftarrow$ \textit{surrogate regression residual} of $X_i$ and $X_j$ on the identified shared latent components $\mathbf{S'}$;
        \IF{$\exists X_k \in \mathbf{V} \setminus \{X_i, X_j\}$,  $(X_k, \{\widetilde{X}_i, \widetilde{X}_j\})$ satisfies the One-Latent-Component condition}
        \STATE Add a new latent confounder $L$ with $X_k \leftarrow L \rightarrow \{X_i, X_j\}$ to $\mathcal{G}$ and update $\widehat{\mathbf{A}}$;
        \ENDIF
    \ENDFOR
    \STATE //Determine the Causal Order
    \FOR{each pairs of variables, $\{X_i, X_j\} \in \mathbf{V}$, not connected by directed edges}
        \STATE $\widetilde{X}_i, \widetilde{X}_j \leftarrow$ \textit{surrogate regression residual} of $X_i$ and $X_j$ on the identified shared latent components $\mathbf{S'}$;
        \IF{$(X_i, \{\widetilde{X}_i, \widetilde{X}_j\})$ satisfies the One-Latent-Component condition}
        \STATE Orient $X_i \to X_j$ in $\mathcal{G}$ and update $\widehat{\mathbf{A}}$;
        \ENDIF
    \ENDFOR
  \UNTIL{$\mathcal{G}$ hasn't changed}
  \STATE Remove redundant edges in $\mathcal{G}$ according to Proposition \ref{pro:purify_parent} and update $\widehat{\mathbf{A}}$.
\end{algorithmic}
\end{algorithm}

\begin{figure}
    \centering
    \begin{tikzpicture}[thick]
        \node[causallatent, xshift= -0.60cm, yshift=  0.50cm] (L1) {$L_1$};
        \node[causallatent, xshift=  0.60cm, yshift=  0.50cm] (L2) {$L_2$};
        \node[causalobs,    xshift= -1.20cm, yshift= -0.50cm] (X1) {$X_1$} ;
        \node[causalobs,    xshift=  0.00cm, yshift= -0.50cm] (X2) {$X_2$} ;
        \node[causalobs,    xshift=  1.20cm, yshift= -0.50cm] (X3) {$X_3$} ;
        \node[causalobs,    xshift= -0.60cm, yshift=  1.50cm] (X4) {$X_4$} ;
        \node[causalobs,    xshift=  0.60cm, yshift=  1.50cm] (X5) {$X_5$} ;
        \node[causaleps,    xshift= -1.20cm, yshift= -1.50cm] (EX1) {$S_{X_1}$} ;
        \node[causaleps,    xshift=  0.00cm, yshift= -1.50cm] (EX2) {$S_{X_2}$} ;
        \node[causaleps,    xshift=  1.20cm, yshift= -1.50cm] (EX3) {$S_{X_3}$} ;
        \node[causaleps,    xshift= -1.80cm, yshift=  1.50cm] (EX4) {$S_{X_4}$} ;
        \node[causaleps,    xshift=  1.80cm, yshift=  1.50cm] (EX5) {$S_{X_5}$} ;
        
        \edge {L1} {X1, X2, X3, X4, X5};
        \edge {L2} {X1, X2, X3, X5};
        \edge {X1} {X2};
        \edge {X2} {X3};
        \edge {EX1} {X1};
        \edge {EX2} {X2};
        \edge {EX3} {X3};
        \edge {EX4} {X4};
        \edge {EX5} {X5};
    \end{tikzpicture}
    \caption{A causal graph with 2 latent variables and 5 observed variables, where $S_{X_i}$ is the exogenous variable of $X_i$.}
  \label{fig:Exp4algorithm}
\end{figure}
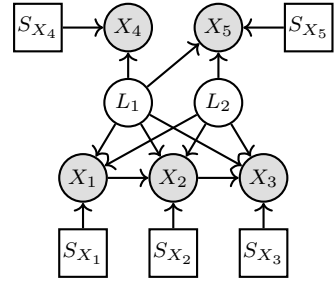

\begin{example}
Considering the example shown in Figure \ref{fig:Exp4algorithm}. We first assume $\{X_1, X_2, X_4\}$ are influenced by $L_1$, then use $X_4$ to estimate the mixing matrix from $L_1$ to $\{X_1, X_2, X_4\}$. Based on the estimated mixing matrix, $(\{X_4\}, \{X_1, X_2\})$ satisfies One-Latent-Component condition, we accept the hypothetical structure with $L_1$. Similarly, we can also identify that $L_1$ affects $\{X_3, X_5\}$. Let $\widetilde{X}_1$ and $\widetilde{X}_2$ be the surrogate regression residuals of $X_1$ and $X_2$ on $L_1$ by utilizing $X_4$ as a surrogate variable, respectively. Then $X_5$ can be used to estimate the mixing matrix from $L_2$ to $\{\widetilde{X}_1, \widetilde{X}_2, X_5\}$. Finally, since $(\{X_5\}, \{\widetilde{X}_1, \widetilde{X}_2\} )$ satisfies the One-Latent-Component condition, we can identify that the latent confounder $L_2$ affects $\{X_1, X_2, X_5\}$. 
\end{example}

\subsubsection{Eliminate the redundant edges}\label{sec:edge_remove}

Now the remaining problem is how to remove the redundant edges from the detected causal order (i.e., finding out whether the observed variables are adjacent to each other). Inspired by the graphical criterion on conditional independence without latent confounders, a variable is independent of its ancestors but non-parent variables, conditioned on its parents. These parents act as mediating variables between the variable and its ancestors. Similarly, if we can find mediate variables and remove their influence on their children, then independent information can be found. That is, the edge between the considered two variables is redundant. The following proposition explains how to find such cases.

\begin{proposition} [Redundant edges elimination] Suppose Assumptions \ref{asm:pure} - \ref{asm:faithful} hold. Let $X_i$ and $X_j$ be two observed variables following the canonical lvLiNGAM model where $X_i$ is causally earlier than $X_j$. There is no directed edge between $X_i$ and $X_j$, if and only if there exists an observed variable set $\mathbf{X_k}$ satisfying that 1) $\mathbf{X_k}$ does not contain  $X_i$ and $X_j$ ; 2) $\forall X_k' \in \mathbf{X_k}$, if $X_k'$ is a descendant of $X_i$, then $X_k'$ must be an ancestor of $X_j$; 3) $\widetilde{X}_j'$ and $\widetilde{X}_j''$ are independent of $X_i$, where $\widetilde{X}_j'$ and $\widetilde{X}_j''$ are the surrogate regression residuals of $X_j$ on the independent components by utilizing $\mathbf{X_k}$ and $\{X_i,\mathbf{X_k} \}$ as surrogates, respectively.
\label{pro:purify_parent}
\end{proposition}

\begin{table*}[t]
  \caption{F1-score (and its variance) of learned directed causal edges with different methods.}
  \label{tab:causal_arrow}
  \centering
\begin{tabular}{cccccccc}
\toprule
\multicolumn{2}{c}{Algorithm} & FCI  & lvLiNGAM & DLiNGAM & PLiNGAM & RCD  & Ours \\
\midrule
\multirow{3}{*}{Case 2} & 500  & 0.00 (0.00) & 0.35 (0.05) & 0.45 (0.02) & 0.20 (0.09) & 0.35 (0.09) & \textbf{0.60 (0.24)} \\
     & 1000 & 0.00 (0.00) & 0.15 (0.05) & 0.50 (0.00) & 0.13 (0.07) & 0.20 (0.09) & \textbf{0.70 (0.21)} \\
     & 2000 & 0.00 (0.00) & 0.30 (0.06) & \textbf{0.50 (0.00)} & 0.07 (0.04) & 0.12 (0.06) & \textbf{0.50 (0.25)} \\
\midrule
\multirow{3}{*}{Case 3} & 500  & 0.00 (0.00) & 0.38 (0.03) & 0.30 (0.03) & 0.43 (0.09) & \textbf{0.44 (0.10)} & 0.35 (0.12) \\
     & 1000 & 0.00 (0.00) & \textbf{0.45 (0.01)} & 0.35 (0.02) & 0.16 (0.06) & 0.22 (0.08) & 0.40 (0.11) \\
     & 2000 & 0.00 (0.00) & 0.43 (0.01) & 0.39 (0.03) & 0.07 (0.04) & 0.13 (0.07) & \textbf{0.47 (0.09)} \\
\midrule
\multirow{3}{*}{Case 4} & 500  & 0.00 (0.00) & 0.31 (0.03) & 0.45 (0.01) & 0.12 (0.06) & 0.21 (0.07) & \textbf{0.58 (0.09)} \\
     & 1000 & 0.00 (0.00) & 0.33 (0.01) & 0.48 (0.01) & 0.05 (0.02) & 0.20 (0.07) & \textbf{0.49 (0.12)} \\
     & 2000 & 0.00 (0.00) & 0.30 (0.02) & \textbf{0.50 (0.00)} & 0.00 (0.00) & 0.04 (0.01) & 0.38 (0.11) \\
\midrule
\multirow{3}{*}{Case 6} & 500  & 0.00 (0.00) & 0.17 (0.02) & 0.29 (0.00) & 0.00 (0.00) & 0.00 (0.00) & \textbf{0.60 (0.24)} \\
     & 1000 & 0.00 (0.00) & 0.17 (0.02) & 0.17 (0.02) & 0.00 (0.00) & 0.00 (0.00) & \textbf{0.40 (0.24)} \\
     & 2000 & 0.00 (0.00) & 0.17 (0.02) & 0.20 (0.02) & 0.00 (0.00) & 0.00 (0.00) & \textbf{0.30 (0.21)} \\
\bottomrule
\end{tabular}%
\end{table*}

Specifically, given two adjacent variables $X_i$ and $X_j$ where $X_i$ is causally earlier than $X_j$ and the identified independent components $\mathbf{S_i}$, we use the following steps to check whether there is a direct edge between them. First, we select the observed variables $\mathbf{X_k}$ that satisfies the condition in Proposition \ref{pro:purify_parent}, to estimate the mixing matrix $\mathbf{A}_{\{X_j, \mathbf{X_k}\}, \mathbf{S_i}}$. Second, if $\widetilde{X}_j'$ is independent of $X_i$, then $X_i$ is not a parent of $X_j$. Note that $\mathbf{S_i}$ is a subset of independent components, which can be latent confounders or shared noises of the observed variables. In practice, we can find $\mathbf{S_i}$ and estimate the corresponding mixing matrix iteratively.

Until now, we can remove the redundant causal edges between observed variables, and determine the causal coefficient matrix based on the estimated mixing matrix and recovered causal structure directly. The identifiability of the causal structure as well as the causal coefficients are summarized in Theorem \ref{th:identify}.

\begin{theorem}
    Suppose Assumptions \ref{asm:pure} - \ref{asm:faithful} hold, and that the data is generated by the canonical lvLiNGAM model. Then the causal structure as well as the causal coefficients can be identified by Algorithm \ref{alg:algorithm_framework}.
\label{th:identify}
\end{theorem}

\subsection{Time Complexity Analysis}

For the computational complexity of the proposed algorithm, the big O runtime of Algorithm \ref{alg:algorithm_framework} is $\mathcal{O}(n^3m)$ in the worst case, where $n$ represents the number of observed variables and $m$ represents the number of latent variables. 

In detail, for the for-loop in lines 4 - 9 of Algorithm \ref{alg:algorithm_framework}, it chooses three observed variables to test One-Latent-Component condition, so the worst computational complexity is $\mathcal{O}(n^3)$. For the for-loop in lines 11 - 16 of Algorithm \ref{alg:algorithm_framework}, it chooses two observed variables to test One-Latent-Component condition, so the worst computational complexity is $\mathcal{O}(n^2)$. According to Lemma \ref{th:ext-one-latent-component}, some latent components need to be identified by iteratively removing the identified latent components. The loop in lines 2 - 17 of Algorithm \ref{alg:algorithm_framework} is to remove the identified latent components iteratively, and the worst computational complexity is $\mathcal{O}(n^3m)$. For line 18 of Algorithm \ref{alg:algorithm_framework}, it chooses two observed variables that are directly connected to remove a redundant edge between them, and the worst computational complexity is $\mathcal{O}(n^2)$.

\section{Experiments}\label{sec:experiments}

\subsection{Synthetic Data}

To mimic real, complex situations, we exploited rather complex simulation settings. In these settings, the observed data is generated according to the lvLiNGAM model, where the causal coefficient $b_{ij}$ is sampled from a uniform distribution between $[0.2, 0.8]$, and the noises are generated from standard normal variables raised to the third power. For each model, the sample size $N$ is varied among $[500, 1000, 2000]$. The details of the causal structures are as follows:

[Case 1]: One latent variable has three observed variables as children and  there is no direct edge among the observed variables, i.e., $L_1 \rightarrow \lbrace X_{1}, X_{2}, X_{3} \rbrace$. 

[Case 2]: Add one edge $X_2 \rightarrow X_3$ to the graph in Case 1. 

[Case 3]: Add one observed variable $X_4$ and three edges $L_1 \rightarrow X_4$, $X_1 \rightarrow X_2$, $X_3 \rightarrow X_4$ to the graph in Case 1.

[Case 4]: Add one observed variable $X_4$ and two edges $L_1 \rightarrow X_4$, $X_3 \rightarrow X_4$ to the graph in Case 2. 

[Case 5]: Add one observed variable $X_4$ and one latent variable $L_2$ that influences three observed variables $\lbrace X_{2}, X_{3}, X_{4}\rbrace$ to the graph in Case 1. 

[Case 6]: Add two edges $X_2 \rightarrow X_3$ and $L_1 \rightarrow X_4$ to the graph in Case 5.

\begin{table*}[t]
  \caption{F1-score (and its variance) of learned non-adjacent relationships with different methods.}
  \label{tab:causal_nonadj}
  \centering
\begin{tabular}{cccccccc}
\toprule
\multicolumn{2}{c}{Algorithm} & FCI  & lvLiNGAM & DLiNGAM & PLiNGAM & RCD  & Ours \\
\midrule
\multirow{3}{*}{Case 1} & 500  & 0.50 (0.00) & 0.00 (0.00) & 0.00 (0.00) & 0.30 (0.06) & 0.35 (0.05) & \textbf{0.92 (0.01)} \\
     & 1000  & 0.35 (0.05) & 0.00 (0.00) & 0.05 (0.02) & 0.15 (0.05) & 0.30 (0.06) & \textbf{0.94 (0.01)} \\
     & 2000  & 0.35 (0.05) & 0.00 (0.00) & 0.00 (0.00) & 0.00 (0.00) & 0.15 (0.05) & \textbf{0.96 (0.01)} \\
\midrule
\multirow{3}{*}{Case 2} & 500  & 0.00 (0.00) & 0.00 (0.00) & 0.00 (0.00) & 0.00 (0.00) & 0.07 (0.04) & \textbf{0.80 (0.16)} \\
     & 1000  & 0.00 (0.00) & 0.00 (0.00) & 0.00 (0.00) & 0.00 (0.00) & 0.00 (0.00) & \textbf{1.00 (0.00)} \\
     & 2000  & 0.00 (0.00) & 0.00 (0.00) & 0.00 (0.00) & 0.00 (0.00) & 0.00 (0.00) & \textbf{0.90 (0.09)} \\
\midrule
\multirow{3}{*}{Case 3} & 500  & 0.74 (0.02) & 0.00 (0.00) & 0.04 (0.01) & 0.39 (0.08) & 0.47 (0.04) & \textbf{0.84 (0.02)} \\
     & 1000  & 0.69 (0.00) & 0.00 (0.00) & 0.00 (0.00) & 0.17 (0.07) & 0.32 (0.08) & \textbf{0.83 (0.03)} \\
     & 2000  & 0.61 (0.01) & 0.00 (0.00) & 0.08 (0.03) & 0.07 (0.04) & 0.32 (0.08) & \textbf{0.88 (0.01)} \\
\midrule
\multirow{3}{*}{Case 4} & 500  & 0.51 (0.02) & 0.04 (0.01) & 0.00 (0.00) & 0.08 (0.03) & 0.16 (0.04) & \textbf{0.88 (0.00)} \\
     & 1000  & 0.48 (0.01) & 0.00 (0.00) & 0.00 (0.00) & 0.04 (0.01) & 0.12 (0.03) & \textbf{0.86 (0.00)} \\
     & 2000  & 0.32 (0.03) & 0.00 (0.00) & 0.00 (0.00) & 0.00 (0.00) & 0.00 (0.00) & \textbf{0.77 (0.07)} \\
\midrule
\multirow{3}{*}{Case 5} & 500  & 0.41 (0.01) & 0.00 (0.00) & 0.00 (0.00) & 0.00 (0.00) & 0.39 (0.05) & \textbf{0.75 (0.01)} \\
     & 1000  & 0.31 (0.00) & 0.00 (0.00) & 0.00 (0.00) & 0.00 (0.00) & 0.33 (0.01) & \textbf{0.80 (0.02)} \\
     & 2000  & 0.29 (0.00) & 0.00 (0.00) & 0.03 (0.01) & 0.00 (0.00) & 0.26 (0.01) & \textbf{0.78 (0.01)} \\
\midrule
\multirow{3}{*}{Case 6} & 500  & 0.44 (0.03) & 0.00 (0.00) & 0.00 (0.00) & 0.00 (0.00) & 0.00 (0.00) & \textbf{0.94 (0.02)} \\
     & 1000  & 0.22 (0.04) & 0.00 (0.00) & 0.00 (0.00) & 0.00 (0.00) & 0.00 (0.00) & \textbf{0.98 (0.01)} \\
     & 2000  & 0.10 (0.02) & 0.00 (0.00) & 0.00 (0.00) & 0.00 (0.00) & 0.00 (0.00) & \textbf{0.98 (0.01)} \\
\bottomrule
\end{tabular}%
\end{table*}

We compare our method with FCI \citep{spirtes2000causation}, lvLiNGAM \citep{hoyer2008estimation}, DLiNGAM \citep{shimizu2011directlingam}, PLiNGAM \citep{tashiro2014parcelingam}, and RCD \citep{maeda2020rcd}. The results are evaluated in terms of directed causal edges between variables and non-adjacent relationships, using Precision, Recall and $F_1$-score as evaluation metrics. In detail, Precision is the percentage of correct directed causal/non-adjacent relationships between observed variables among all directed causal/non-adjacent relationships returned by the algorithm. Recall is the percentage of correct directed causal/non-adjacent relationships that are found by the search among true directed causal/non-adjacent relationships between observed variables. $F_1$-score is defined as $F_1 = \frac{2\times Precision \times Recall}{Precision + Recall}$. Besides, we use Root Mean Square Errors (RMSE) as a metric to evaluate the performance of causal coefficient estimation and compute the average computation time to show the efficiency of our method. Each experiment was repeated 10 times with randomly generated data and the results were averaged. Precision and Recall, RMSE and average computation time of all methods as well as more experimental results are given in Appendix \ref{app:exp_six_cases}. We also conducted experiments in a more simple, easy-to-learn setting and three settings on violating assumptions of our method, and the corresponding experimental results are provided in Appendix \ref{app:exp_syn} - \ref{app:exp_asm}, respectively.

\textbf{Evaluation on directed causal edges.}
In Table \ref{tab:causal_arrow}, we only show the results of Cases 2 - 4, and Case 6, because Case 1 and Case 5 are no directed edges between observed variables in the truth graph. The value in parentheses indicates the variance of F1-score. As shown in Table \ref{tab:causal_arrow}, most methods obtain low F1-scores. This is because the considered settings are harder to learn. The number of directed edges in the true graph is very small, so the F1-scores is very low if any causal edges are incorrectly learned. From the result, our algorithm performs well in almost all cases. This implies that the pure set of observed variables helps to determine the causal relations. lvLiNGAM obtains a lower F1-score than ours except Case 3. Because its performance depends on the estimation of the mixing matrix and the transformation from the mixing matrix to the causal coefficient matrix, even when the mixing matrix estimated by OICA is correct, the causal coefficient matrix also relies on which columns of the mixing matrix correspond to the latent variables, which leads to the low $F_1$-scores. FCI obtains the worst results. Because in these cases, all variables are influenced by the latent confounder, which makes FCI return undetermined edges. The other methods without the pure set of observed variables as a surrogate variable of latent confounder, falsely determine the spurious relations between observed variables. Considering the variance of F1-score, our method performs worst in case 2 and case 6. This is because there is only one true edge in the graph and the result is either all right or none (i.e., F1-score is 1 or 0). 

\textbf{Evaluation on non-adjacent relationships.} As shown in Table \ref{tab:causal_nonadj}, our algorithm achieves the highest precision, recall, and $F_1$-score on all cases. FCI achieves the second-best performance in Cases 3 - 5, which implies that the independence-based method can remove some edges between two variables that are not shared by the same latent confounders. But it performs worst in Case 2 and Case 6. PLiNGAM performs poorly in recovering the non-adjacent relationships in all cases, because it cannot determine the causal relations between two observed variables that share the same latent confounders. RCD can remove a few causal edges in Case 5 because $X_1$ and $X_4$ are independent without conditioning on the latent confounders. DLiNGAM also perform poorly due to the unsatisfied causal sufficient assumption. lvLiNGAM obtains the worst result, and cannot discover any non-adjacent relationships, because it is based on model selection and prefers a dense graph without a sparse constraint. Considering the variance of F1-score, the variance of many methods is zero, because the mean of their result is zero. In these cases, our method can obtain much higher means and slightly small variances, which reflects the reliability and soundness in learning non-adjacent relationships.

\subsection{Stock Market Data}
We applied our method to the Hong Kong Stock Market Data, aiming at discovering the causal relationships among the selected 14 stocks. The data contains 1331 daily returns. More details are provided in Appendix \ref{app:stock}.

The found causal relationships are: 

1) all observed variables are affected by a latent confounder. Besides, all observed variables except $X_7$ and $X_{14}$ are influenced by another latent confounder at the same time; 

2) the directed causal edges between observed variables are: $X_2 \to \{X_3, X_8 \}$, $X_9 \to X_4$, $X_{11} \to X_{10} \to X_1$; 3) the undirected edges are $X_1 - X_4$, $X_2 -X_6$, $X_1 -  X_{13} - X_{14}$. 

The estimated causal structure inspires us with some findings as follows: 

1) the whole market environment is affected by the hidden common factors (which may be policy, the total risk in the market and so on) \citep{cai2019triad}. This is also consistent with the expert knowledge of the stock market. 

2) There are often connections among stocks belonging to the same sub-index. For example, $X_2$, $X_3$ and $X_6$ are dependent because they are constituent stocks under the Hang Seng Utilities Index. 

3) The edge $X_{10} \to X_1$ is consistent with the knowledge that ownership relations ($X_1$ holds about 50\% of $X_{10}$) tend to cause causal relations \citep{zhang2008minimal}. 

\section{Discussion}\label{sec:disscution}

In practice, the causal graph may be very complex, as seen from the extensive studies provided in Appendix \ref{app:exp_random}, and our assumptions may be violated in some way. So in this section, we discuss the limitations of the proposed procedure in real-world applications. 

The validity of our assumptions generally depends on the domain. If the number of latent variables, relative to the observed variables, is not very large, our assumptions may hold true in many problems. However, please notice that if our assumptions do not hold true, the output of the procedure may indicate it, to avoid misleading results in complex scenarios. Theoretically, the contrapositive of Theorem \ref{th:identify} implies that if the causal structure produced by our Algorithm \ref{alg:algorithm_framework} contains undirected links, then the data violate at least one of the three assumptions somewhere (or it is because of the finite sample size effect). On the other hand, if one observes such a complete undirected sub-graph in the output, it strongly suggests violations of certain assumptions. One may conclude that certain assumptions are violated, or try to use other methods to go further.

Besides, we also conducted experiments on synthesis data to evaluate the behavior of the procedure when various assumptions are violated; empirical results are given in Appendix \ref{app:exp_asm}.

Furthermore, our correctness result assumes a large sample size, and in practice, the result may be sensitive to random errors on finite samples. Thus, we have emphasized that one should pay more attention to the assumptions of our procedure and try to validate the results before using it.

\section{Conclusion}\label{sec:conclusion}
We investigated the causal discovery problem with latent confounders by using higher-order cumulants. First, we proposed a closed-form solution to OICA in specific cases, which explicitly estimates the mixing matrix using the higher-order cumulants. Second, we further extended the results to estimate a canonical lvLiNGAM, by iteratively employing the proposed One-Latent-Component condition to test for the existence of latent confounders and determine the causal directions between observed variables. Experimental results further verified the correctness and effectiveness of our algorithm. Future research along this line includes allowing nonlinear causal relationships and relaxing the independence assumption among the latent variables.

\section*{Acknowledgements}
This research was supported in part by National Key R\&D Program of China (2021ZD0111501), National Science Fund for Excellent Young Scholars (62122022), Natural Science Foundation of China (61876043, 61976052, 62206064), the major key project of PCL (PCL2021A12). KZ was supported in part by the NSF-Convergence Accelerator Track-D award \#2134901, by the National Institutes of Health (NIH) under Contract R01HL159805, by grants from Apple Inc., KDDI Research, Quris AI, and IBT, and by generous gifts from Amazon, Microsoft Research, and Salesforce. We sincerely appreciate the insightful discussions from Zhouchen Lin, Yang Chen and anonymous reviewers, which greatly helped to improve the paper.

\bibliography{icml2023}

\begin{thebibliography}{31}
\providecommand{\natexlab}[1]{#1}
\providecommand{\url}[1]{\texttt{#1}}
\expandafter\ifx\csname urlstyle\endcsname\relax
  \providecommand{\doi}[1]{doi: #1}\else
  \providecommand{\doi}{doi: \begingroup \urlstyle{rm}\Url}\fi

\bibitem[Adams et~al.(2021)Adams, Hansen, and Zhang]{adams2021identification}
Adams, J., Hansen, N., and Zhang, K.
\newblock Identification of partially observed linear causal models: Graphical
  conditions for the non-gaussian and heterogeneous cases.
\newblock In \emph{Advances in Neural Information Processing Systems},
  volume~34, pp.\  22822--22833, 2021.

\bibitem[Brillinger(2001)]{brillinger2001time}
Brillinger, D.~R.
\newblock \emph{Time series: data analysis and theory}.
\newblock Society for Industrial and Applied Mathematics, 2001.

\bibitem[Cai et~al.(2019)Cai, Xie, Glymour, Hao, and Zhang]{cai2019triad}
Cai, R., Xie, F., Glymour, C., Hao, Z., and Zhang, K.
\newblock Triad constraints for learning causal structure of latent variables.
\newblock \emph{Advances in Neural Information Processing Systems}, 2019.

\bibitem[Chen et~al.(2021)Chen, Cai, Zhang, and Hao]{chen2021causal}
Chen, W., Cai, R., Zhang, K., and Hao, Z.
\newblock Causal discovery in linear non-gaussian acyclic model with multiple
  latent confounders.
\newblock \emph{IEEE Transactions on Neural Networks and Learning Systems},
  2021.

\bibitem[Colombo \& Maathuis(2014)Colombo and Maathuis]{colombo2014order}
Colombo, D. and Maathuis, M.~H.
\newblock Order-independent constraint-based causal structure learning.
\newblock \emph{Journal of Machine Learning Research}, 15\penalty0
  (116):\penalty0 3921--3962, 2014.

\bibitem[Colombo et~al.(2012)Colombo, Maathuis, Kalisch, and
  Richardson]{colombo2012learning}
Colombo, D., Maathuis, M.~H., Kalisch, M., and Richardson, T.~S.
\newblock Learning high-dimensional directed acyclic graphs with latent and
  selection variables.
\newblock \emph{The Annals of Statistics}, pp.\  294--321, 2012.

\bibitem[Comon \& Jutten(2010)Comon and Jutten]{comon2010handbook}
Comon, P. and Jutten, C.
\newblock \emph{Handbook of Blind Source Separation: Independent component
  analysis and applications}.
\newblock Academic press, 2010.

\bibitem[Darmois(1953)]{darmois1953analyse}
Darmois, G.
\newblock Analyse g{\'e}n{\'e}rale des liaisons stochastiques: etude
  particuli{\`e}re de l'analyse factorielle lin{\'e}aire.
\newblock \emph{Revue de l'Institut international de statistique}, pp.\  2--8,
  1953.

\bibitem[Eriksson \& Koivunen(2004)Eriksson and
  Koivunen]{eriksson2004identifiability}
Eriksson, J. and Koivunen, V.
\newblock Identifiability, separability, and uniqueness of linear ica models.
\newblock \emph{IEEE Signal Processing Letters}, 11\penalty0 (7):\penalty0
  601--604, 2004.

\bibitem[Feller(1991)]{feller1991introduction}
Feller, W.
\newblock \emph{An introduction to probability theory and its applications,
  Volume 2}, volume~81.
\newblock John Wiley \& Sons, 1991.

\bibitem[H{\o}jen-S{\o}rensen et~al.(2002)H{\o}jen-S{\o}rensen, Winther, and
  Hansen]{hojen2002mean}
H{\o}jen-S{\o}rensen, P.~A., Winther, O., and Hansen, L.~K.
\newblock Mean-field approaches to independent component analysis.
\newblock \emph{Neural Computation}, 14\penalty0 (4):\penalty0 889--918, 2002.

\bibitem[Hoyer et~al.(2008)Hoyer, Shimizu, Kerminen, and
  Palviainen]{hoyer2008estimation}
Hoyer, P.~O., Shimizu, S., Kerminen, A.~J., and Palviainen, M.
\newblock Estimation of causal effects using linear non-gaussian causal models
  with hidden variables.
\newblock \emph{International Journal of Approximate Reasoning}, 49\penalty0
  (2):\penalty0 362--378, 2008.

\bibitem[Hyv{\"a}rinen \& Smith(2013)Hyv{\"a}rinen and
  Smith]{hyvarinen2013pairwise}
Hyv{\"a}rinen, A. and Smith, S.~M.
\newblock Pairwise likelihood ratios for estimation of non-gaussian structural
  equation models.
\newblock \emph{Journal of Machine Learning Research}, 14\penalty0
  (Jan):\penalty0 111--152, 2013.

\bibitem[Hyv{\"a}rinen et~al.(2001)Hyv{\"a}rinen, Karhunen, and
  Oja]{hyvarinen2001independent}
Hyv{\"a}rinen, A., Karhunen, J., and Oja, E.
\newblock Independent component analysis, adaptive and learning systems for
  signal processing, communications, and control.
\newblock \emph{John Wiley \& Sons, Inc}, 1:\penalty0 11--14, 2001.

\bibitem[Krasilnikov et~al.(2019)Krasilnikov, Beregun, and
  Harmash]{krasilnikov2019analysis}
Krasilnikov, A., Beregun, V., and Harmash, O.
\newblock Analysis of estimation errors of the fifth and sixth order cumulants.
\newblock In \emph{International Conference on Electronics and Nanotechnology
  (ELNANO)}, pp.\  754--759. IEEE, 2019.

\bibitem[Lewicki \& Sejnowski(2000)Lewicki and Sejnowski]{lewicki2000learning}
Lewicki, M.~S. and Sejnowski, T.~J.
\newblock Learning overcomplete representations.
\newblock \emph{Neural Computation}, 12\penalty0 (2):\penalty0 337--365, 2000.

\bibitem[Maeda \& Shimizu(2020)Maeda and Shimizu]{maeda2020rcd}
Maeda, T.~N. and Shimizu, S.
\newblock Rcd: Repetitive causal discovery of linear non-gaussian acyclic
  models with latent confounders.
\newblock In \emph{International Conference on Artificial Intelligence and
  Statistics}, pp.\  735--745. PMLR, 2020.

\bibitem[Olshausen \& Millman(1999)Olshausen and
  Millman]{olshausen1999learning}
Olshausen, B. and Millman, K.
\newblock Learning sparse codes with a mixture-of-gaussians prior.
\newblock \emph{Advances in Neural Information Processing Systems}, 12, 1999.

\bibitem[Pandav et~al.(2019)Pandav, Mallick, and Mohanty]{pandav2019effect}
Pandav, A., Mallick, D., and Mohanty, B.
\newblock Effect of limited statistics on higher order cumulants measurement in
  heavy-ion collision experiments.
\newblock \emph{Nuclear Physics A}, 991:\penalty0 121608, 2019.

\bibitem[Salehkaleybar et~al.(2020)Salehkaleybar, Ghassami, Kiyavash, and
  Zhang]{salehkaleybar2020learning}
Salehkaleybar, S., Ghassami, A., Kiyavash, N., and Zhang, K.
\newblock Learning linear non-gaussian causal models in the presence of latent
  variables.
\newblock \emph{Journal of Machine Learning Research}, 21:\penalty0 39--1,
  2020.

\bibitem[Schefczik \& Hägele(2019)Schefczik and Hägele]{schefczik2019ready}
Schefczik, F. and Hägele, D.
\newblock Ready-to-use unbiased estimators for multivariate cumulants including
  one that outperforms $\overline{x^3}$.
\newblock \emph{arXiv preprint arXiv:1904.12154}, 2019.

\bibitem[Shimizu et~al.(2011)Shimizu, Inazumi, Sogawa, Hyv{\"a}rinen, Kawahara,
  Washio, Hoyer, and Bollen]{shimizu2011directlingam}
Shimizu, S., Inazumi, T., Sogawa, Y., Hyv{\"a}rinen, A., Kawahara, Y., Washio,
  T., Hoyer, P.~O., and Bollen, K.
\newblock Directlingam: A direct method for learning a linear non-gaussian
  structural equation model.
\newblock \emph{Journal of Machine Learning Research}, 12:\penalty0 1225--1248,
  2011.

\bibitem[Silva et~al.(2006)Silva, Scheines, Glymour, Spirtes, and
  Chickering]{silva2006learning}
Silva, R., Scheines, R., Glymour, C., Spirtes, P., and Chickering, D.~M.
\newblock Learning the structure of linear latent variable models.
\newblock \emph{Journal of Machine Learning Research}, 7\penalty0 (2), 2006.

\bibitem[Skitovitch(1953)]{skitovitch1953property}
Skitovitch, V.
\newblock On a property of the normal distribution.
\newblock \emph{DAN SSSR}, 89:\penalty0 217--219, 1953.

\bibitem[Spirtes et~al.(1995)Spirtes, Meek, and Richardson]{spirtes1995causal}
Spirtes, P., Meek, C., and Richardson, T.
\newblock Causal inference in the presence of latent variables and selection
  bias.
\newblock In \emph{Conference on Uncertainty in artificial intelligence}, pp.\
  499--506, 1995.

\bibitem[Spirtes et~al.(2000)Spirtes, Glymour, Scheines, and
  Heckerman]{spirtes2000causation}
Spirtes, P., Glymour, C.~N., Scheines, R., and Heckerman, D.
\newblock \emph{Causation, prediction, and search}.
\newblock MIT press, 2000.

\bibitem[Tashiro et~al.(2014)Tashiro, Shimizu, Hyv{\"a}rinen, and
  Washio]{tashiro2014parcelingam}
Tashiro, T., Shimizu, S., Hyv{\"a}rinen, A., and Washio, T.
\newblock Parcelingam: a causal ordering method robust against latent
  confounders.
\newblock \emph{Neural Computation}, 26\penalty0 (1):\penalty0 57--83, 2014.

\bibitem[Wang \& Drton(2020{\natexlab{a}})Wang and Drton]{wang2020causal}
Wang, Y.~S. and Drton, M.
\newblock Causal discovery with unobserved confounding and non-gaussian data.
\newblock \emph{arXiv preprint arXiv:2007.11131}, 2020{\natexlab{a}}.

\bibitem[Wang \& Drton(2020{\natexlab{b}})Wang and Drton]{wang2020high}
Wang, Y.~S. and Drton, M.
\newblock High-dimensional causal discovery under non-gaussianity.
\newblock \emph{Biometrika}, 107\penalty0 (1):\penalty0 41--59,
  2020{\natexlab{b}}.

\bibitem[Xie et~al.(2020)Xie, Cai, Huang, Glymour, Hao, and
  Zhang]{xie2020generalized}
Xie, F., Cai, R., Huang, B., Glymour, C., Hao, Z., and Zhang, K.
\newblock Generalized independent noise condition for estimating latent
  variable causal graphs.
\newblock In \emph{Advances in Neural Information Processing Systems}, 2020.

\bibitem[Zhang \& Chan(2008)Zhang and Chan]{zhang2008minimal}
Zhang, K. and Chan, L.
\newblock Minimal nonlinear distortion principle for nonlinear independent
  component analysis.
\newblock \emph{Journal of Machine Learning Research}, 9\penalty0
  (Nov):\penalty0 2455--2487, 2008.

\end{thebibliography}
\bibliographystyle{icml2023}

%%%%%%%%%%%%%%%%%%%%%%%%%%%%%%%%%%%%%%%%%%%%%%%%%%%%%%%%%%%%%%%%%%%%%%%%%%%%%%%
%%%%%%%%%%%%%%%%%%%%%%%%%%%%%%%%%%%%%%%%%%%%%%%%%%%%%%%%%%%%%%%%%%%%%%%%%%%%%%%
% APPENDIX
%%%%%%%%%%%%%%%%%%%%%%%%%%%%%%%%%%%%%%%%%%%%%%%%%%%%%%%%%%%%%%%%%%%%%%%%%%%%%%%
%%%%%%%%%%%%%%%%%%%%%%%%%%%%%%%%%%%%%%%%%%%%%%%%%%%%%%%%%%%%%%%%%%%%%%%%%%%%%%%
\newpage
\appendix
\onecolumn

\section*{Appendix}

In App.\ref{app:proof}, we provide the proof of theorems, corollaries, and propositions. In App.\ref{app:est_by_general_cum}, we provide a detailed procedure to use the general higher-order cumulants. In App.\ref{app:toy}, we provide a running example of our proposed algorithm. In App.\ref{app:cumest}, we provide the estimating variance of the higher-order statistics. In App.\ref{app:exp}, we provide more details of experiments and theoretical analysis.

\section{Proofs}\label{app:proof}

Before providing the proofs, we need to introduce the following important theorem first.

\begin{theorem}
    [Darmois-Skitovitch Theorem] \citep{darmois1953analyse, skitovitch1953property}. Define two random variables $X_{1}$ and $X_{2}$, as linear combinations of independent random variables $S_i, i=1, \dots, n$:
	\begin{equation}
	X_{1} = \sum_{i=1}^{n}\alpha_{i}S_{i},
	X_{2} = \sum_{i=1}^{n}\beta_{i}S_{i}.
	\end{equation}
	
	If $X_{1}$ and $X_{2}$ are statistically independent, then all variables $S_j$ for which $\alpha_{j}\beta_{j} \neq 0$ are Gaussian.
	\label{th:ds}
\end{theorem}

In other words, if random variables $S_i, i=1, \dots, n$ are independent and for some $\alpha_{1}, \alpha_{2},\dots, \alpha_{n}$ and $\beta_{1}, \beta_{2},\dots, \beta_{n}$, $X_1$ is independent of $X_2$, then for any $S_{j}$ that is non-Gaussian, at most one of $\alpha_{j}$ and $\beta_{j}$ can be nonzero.

\subsection{Proof of Theorem \ref{th:shared-one-exogenous}}\label{app:coef_estimation}

\textbf{Theorem \ref{th:shared-one-exogenous}}. Let $X_i$ and $X_j$ be two observed variables following Eq. (\ref{eq:ica}). Suppose $X_i$ and $X_j$ follow the One-Latent-Component structure and that $S$ is the only one shared non-Gaussian latent component of them and has a unit variance. Then the mixing coefficients between $\{X_i, X_j\}$ and $S$, denoted by $\hat{\alpha}_i$ and $\hat{\alpha}_j$ respectively, can be identified by the fourth-order cumulant as follows:
\begin{equation}
    \begin{aligned}
        \hat{\alpha}_i &= \sqrt{\frac{cum(X_i, X_i, X_j, X_j)}{cum(X_i, X_j, X_j, X_j)} \cdot cum(X_i, X_j)}, \\
                     % &= \sqrt{\frac{cum(X_i, X_i, X_i, X_j)}{cum(X_i, X_i, X_j, X_j)} \cdot cum(X_i, X_j)},\\
        \hat{\alpha}_j &= \frac{cum(X_i, X_j)}{\hat{\alpha}_i}. 
    \end{aligned}
\end{equation}

\textit{Proof}. Suppose $X_i$ and $X_j$ follow the One-Latent-Component structure, and $S$ is the only one shared latent component of them. The generating process of $X_i$ and $X_j$ can be written in terms of the mixing matrix:
\begin{equation}
    \begin{aligned}
        X_i &= \alpha_i S + \mathbf{A}_{X_i, \mathbf{S_i'}}\mathbf{S_i'},\\
        X_j &= \alpha_j S + \mathbf{A}_{X_j, \mathbf{S_j'}}\mathbf{S_j'},
    \end{aligned}
\end{equation}
where $\alpha_i$ and $\alpha_j$ are the mixing coefficients between $\{X_i, X_j\}$ and $S$, respectively. $S$, $\mathbf{S_i'}$ and $\mathbf{S_j'}$ are independent of each other.
To obtain the mixing coefficients $\alpha$ and $\beta$, we consider the cumulant of $(X_i, X_j)$, $(X_i, X_i, X_j, X_j)$ and $(X_i, X_j, X_j, X_j)$  as:
\begin{equation}
    \begin{aligned}
        cum(X_i, X_j) &= \alpha_i \alpha_j cum(S, S),\\
        cum(X_i, X_i, X_j, X_j) & =\alpha_i^{2} \alpha_j^{2} cum(S, S, S, S),\\
        cum(X_i, X_j, X_j, X_j) & =\alpha_i \alpha_j^{3} cum(S, S, S, S).\\
    \end{aligned}
\end{equation}
 
Then the square of $\alpha_i$ is obtained in the following way:
\begin{equation}
    \begin{aligned}
          & \frac{cum(X_i, X_i, X_j, X_j)}{cum(X_i, X_j, X_j, X_j)} \cdot cum(X_i, X_j) \\
        = &\frac{\alpha_i}{\alpha_j} \cdot (\alpha_i \alpha_j cum(S, S)) \\
        =&\frac{\alpha_i}{\alpha_j} \cdot \alpha_i \alpha_j \sigma_{S}^{2} \\
        =& \alpha_i^2 \sigma_{S}^2,
    \end{aligned}
    \label{eq:mixing_coef1}
\end{equation}
where $\sigma_{S}$ is the standard deviation of the non-Gaussian exogenous variable $S$. Similarly, we have the square of $\alpha_j$ as:
\begin{equation}
    \begin{aligned}
          & \frac{cum(X_i, X_i, X_j, X_j)}{cum(X_i, X_i, X_i, X_j)} \cdot cum(X_i, X_j) \\
        = &\frac{\alpha_j}{\alpha_i} \cdot (\alpha_i \alpha_j cum(S, S)) \\
        = & \frac{\alpha_j}{\alpha_i} \cdot \alpha_i \alpha_j \sigma_{S}^{2} \\
        = & \alpha_j^2 \sigma_{S}^2.
    \end{aligned}
    \label{eq:mixing_coef2}
\end{equation}
Assumed that $S$ has a unit variance, the square of the mixing coefficient can be estimated directly from Eq. (\ref{eq:mixing_coef1}) or Eq. (\ref{eq:mixing_coef2}).

To determine the sign of the mixing coefficients, we can use the cumulant of $(X_i,X_j)$ as:
\begin{equation}
    \begin{aligned}
        \hat{\alpha}_i &= \sqrt{\frac{cum(X_i, X_i, X_j, X_j)}{cum(X_i, X_j, X_j, X_j)} \cdot cum(X_i, X_j)},\\
        \hat{\alpha}_j  &= \frac{cum(X_i, X_j)}{\hat{\alpha}}. \\
    \end{aligned}
    \label{eq:mixing_coef}
\end{equation}
Thus, $\alpha_i$ and $\alpha_j$ can be estimated by the cumulants according to Eq. (\ref{eq:mixing_coef}). \qed

\subsection{Proof of Theorem \ref{th:shared-multiple-exogenous}}

\textbf{Theorem \ref{th:shared-multiple-exogenous}}. Let $X_i$ and $X_j$ be two observed variables and $\mathbf{X_k}$ be an observed variable set following Eq. (\ref{eq:ica}) (where $X_i, X_j \notin \mathbf{X_k}$), $\mathbf{S_1}$ be an independent component set and $S_2$ be an independent component not in $\mathbf{S_1}$. Suppose $\{\mathbf{S_{1}},  S_{2} \}$ are the shared independent components of $X_i$ and $X_j$, and that $\mathbf{S_{1}}$ are the shared independent components of $\mathbf{X_k}$ and $\lbrace X_i, X_j \rbrace$. Then the mixing coefficient from $S_2$ to $\{X_i, X_j\}$ can be identified when $X_i$ and $\widetilde{X}_j$ share only one independent component $S_2$, where $\widetilde{X}_j$ is the \emph{surrogate regression residual} of $X_j$ on $\mathbf{S_1}$ by utilizing $\mathbf{X_k}$ as surrogates.

\textit{Proof}. Let $X_i$ and $X_j$ be two observed variable and $\mathbf{X_k}$ be an observed variable set following Eq. (\ref{eq:ica}) (where $X_i, X_j \notin \mathbf{X_k}$), $\mathbf{S_1}$ be an independent component set and $S_2$ be an independent component not in $\mathbf{S_1}$. Suppose $\{\mathbf{S_{1}},  S_{2} \}$ are the shared independent components of $X_i$ and $X_j$, and $\mathbf{S_{1}}$ are the shared independent components of $\mathbf{X_k}$ and $\lbrace X_i, X_j \rbrace$. We can rewrite the generation process of $\{X_i, X_j, \mathbf{X_k}\}$ as follows:
\begin{equation}
    \begin{aligned}
    X_i &= \mathbf{A}_{X_i,\mathbf{S_1}}\mathbf{S_1} + \mathbf{A}_{X_i, S_2}S_2 + S_{i}',\\
    X_j &= \mathbf{A}_{X_j,\mathbf{S_1}}\mathbf{S_1} + \mathbf{A}_{X_j, S_2}S_2 + S_{j}',\\
    \mathbf{X_k} &= \mathbf{A}_{\mathbf{X_k},\mathbf{S_1}}\mathbf{S_1} + \mathbf{A}_{\mathbf{X_k},\mathbf{S_{k}'}}\mathbf{S_{k}'},
    \end{aligned}
\end{equation}
where $S_{i}', S_{j}', \mathbf{S_{k}'}$ are mutually independent. Then $\widetilde{X}_j$ shares the only independent component $S_2$ with observed variable $X_j$. We can rewrite $\widetilde{X}_j$ as follows:
\begin{equation}
    \begin{aligned}
    \widetilde{X}_j
    &= 
    \omega^\top
    \begin{bmatrix}
        X_j\\
        \mathbf{X_k}
    \end{bmatrix}
    \\
    &=
    \omega^\top
    \begin{bmatrix}
        \mathbf{A}_{X_j,\mathbf{S_1}}\\
        \mathbf{A}_{\mathbf{X_k},\mathbf{S_1}}
    \end{bmatrix}
    \mathbf{S_1}
    +
    \omega^\top
    \begin{bmatrix}
        \mathbf{A}_{X_j, S_2} & 1 & \mathbf{0}\\
        0 & 0 & \mathbf{A}_{\mathbf{X_k},\mathbf{S_{k}'}}\\
    \end{bmatrix}
    \begin{bmatrix}
        S_2\\
        S_{j}'\\
        \mathbf{S_{k}'}\\
    \end{bmatrix}\\
    &= 
    \omega^\top
    \begin{bmatrix}
        \mathbf{A}_{X, S_2} & 1 & \mathbf{0}\\
        0 & 0 & \mathbf{A}_{\mathbf{X_k},\mathbf{S_{k}'}}\\
    \end{bmatrix}
    \begin{bmatrix}
        S_2\\
        S_{i}'\\
        \mathbf{S_{k}'}\\
    \end{bmatrix},\\
    \end{aligned}
\end{equation}
where $\omega^\top\begin{bmatrix} \mathbf{A}_{X_j,\mathbf{S_1}}\\\mathbf{A}_{\mathbf{X_k},\mathbf{S_1}}\end{bmatrix} =\mathbf{0}$. When $Dim(\{X_j, \mathbf{X_k}\}) > Rank\left(\begin{bmatrix} \mathbf{A}_{X_j,\mathbf{S_1}}\\\mathbf{A}_{\mathbf{X_k},\mathbf{S_1}}\end{bmatrix}\right)$, $\omega$ has infinite solutions. We can choose a solution satisfying $\omega_{X_j}=1$. This allows the mixing coefficient between the shared independent component $S_2$ and $\widetilde{X}_j$ to be consistent with the mixing coefficient between the shared independent component $S_2$ and the original variable $X_j$. Thus, we can transform the estimation of the mixing coefficients between $\{X_i, X_j\}$ and $S_2$ into the estimation of the mixing coefficients between $\{X_i, \widetilde{X}_j\}$ and $S_2$. \qed

\subsection{Proof of Theorem \ref{th:condition}}

\textbf{Theorem \ref{th:condition}.} Suppose Assumptions \ref{asm:pure} - \ref{asm:faithful} hold. Let $\mathbf{X_i}$ and $\mathbf{X_j}$ be two dependent subsets of observed variables following the canonical lvLiNGAM model. $(\mathbf{X_i}, \mathbf{X_j})$ satisfies One-Latent-Component condition if and only if $\mathbf{X_i}$ and $\mathbf{X_j}$ follow the One-Latent-Component structure.

\textit{Proof}. 
Let $\mathbf{X_i}$ shares independent components $\mathbf{S_s'}$ with $\mathbf{X_j}$. We can rewrite the generation process of $\mathbf{X_i}$ and $\mathbf{X_j}$ as follows: 
\begin{equation}
    \begin{bmatrix}
    \mathbf{X_i}\\
    \mathbf{X_j}
    \end{bmatrix}
    =
    \begin{bmatrix}
    \mathbf{A}_{\mathbf{X_i},\mathbf{S_s'}}\\
    \mathbf{A}_{\mathbf{X_j},\mathbf{S_s'}}
    \end{bmatrix}
    \mathbf{S_s'}
    +
    \begin{bmatrix}
    \mathbf{A}_{\mathbf{X_i},\mathbf{S_i'}} & \mathbf{0}\\
    \mathbf{0} & \mathbf{A}_{\mathbf{X_j},\mathbf{S_j'}}
    \end{bmatrix}
    \begin{bmatrix}
    \mathbf{S_i'}\\
    \mathbf{S_j'}
    \end{bmatrix},
    \label{eq:shared_components}
\end{equation}
where $\mathbf{S_s'}$, $\mathbf{S_i'}$ and $\mathbf{S_j'}$ are mutually independent. 

The ``if" part: If $\mathbf{X_i}$ and $\mathbf{X_j}$ follow the One-Latent-Component structure, we can estimate the mixing coefficients $\mathbf{\widehat{A}}_{\mathbf{X_i}, L}$ and $\mathbf{\widehat{A}}_{\mathbf{X_j}, L}$ consistent with the generation process according to Theorem \ref{th:shared-one-exogenous}. Because the dimension of $\mathbf{X_j}$ is greater than 2, the dimension of $\mathbf{X_j}$ is greater than the rank of $\mathbf{\widehat{A}}_{\mathbf{X_j}, L}$. Hence, there exists non-zero $\omega$ such that $\omega^{\top}\mathbf{A}_{\mathbf{X_j},L} =0$ and
\begin{equation}
    \omega^{\top}\mathbf{X_j} = \omega^{\top}\mathbf{A}_{\mathbf{X_j},\mathbf{S_s'}} + \omega^{\top}\mathbf{A}_{\mathbf{X_j}, \mathbf{S_j'}}
    =\omega^{\top}\mathbf{A}_{\mathbf{X_j}, \mathbf{S_j'}}.
\end{equation}
We can further simplify the generation process as $\mathbf{S_i'} \rightarrow \mathbf{Z} \leftarrow L, \omega^{\top}\mathbf{X_j} \leftarrow \mathbf{S_j'}$. From the above analysis, we can conclude that $\omega^{\top}\mathbf{X_j}$ is independent of $\mathbf{X_i}$, i.e., $(\mathbf{X_i}, \mathbf{X_j})$ satisfies the One-Latent-Component condition.

The ``only if" part: Suppose $(\mathbf{X_i}, \mathbf{X_j})$ satisfies the One-Latent-Component condition. There must exist non-zero $\omega$ satisfies $\omega^\top\mathbf{A}_{\mathbf{X_j}, \mathbf{S_s'}} = \mathbf{0}$ and $\omega \neq \mathbf{0}$ such that $\omega^\top\mathbf{X_j} \ci \mathbf{X_i}$. If $\mathbf{X_i}$ and $\mathbf{X_j}$ do not follow the One-Latent-Component structure, i.e., $\mathbf{X_i}$ and $\mathbf{X_j}$ share multiple independent components. Under the faithfulness assumptions, $\omega^\top\mathbf{\widehat{A}}_{\mathbf{X_j}, L}=\mathbf{0}$ does not guarantee that $\omega^\top\mathbf{A}_{\mathbf{X_j}, \mathbf{S_s'}}=\mathbf{0}$ according to Theorem \ref{th:shared-one-exogenous}. Hence, $\mathbf{X_i}$ and $\mathbf{X_j}$ follow the One-Latent-Component structure. \qed

\subsection{Proof of Corollary \ref{pro:non-descendant}}

\textbf{Corollary \ref{pro:non-descendant}} Suppose Assumptions \ref{asm:pure} - \ref{asm:faithful} hold. Let $\mathbf{X_i}$ and $\mathbf{X_j}$ be disjoint and dependent subsets of the observed variables following the canonical lvLiNGAM model. If $(\mathbf{X_i}, \mathbf{X_j})$ satisfies the One-Latent-Component condition, then $\mathbf{X_i}$ and $\mathbf{X_j}$ are directly caused by one latent confounder and there is no directed path between $\mathbf{X_i}$ and $\mathbf{X_j}$.

\textit{Proof}. We will prove it by contradiction. Supposed that $\mathbf{X_i}$ and $\mathbf{X_j}$ are affected by more than one latent variable. If $\mathbf{X_i}$ and $\mathbf{X_j}$ are affected by two or more latent variables or have more than one directed path between $\mathbf{X_i}$ and $\mathbf{X_j}$. Then they must share two or more independent components. According to Theorem \ref{th:condition}, $(\mathbf{X_i}, \mathbf{X_j})$ won't satisfy the One-Latent-Component condition. \qed

% The ``only if" part: If $\mathbf{X_i}$ and $\mathbf{X_j}$ are affected by the same latent variable and there is no directed path between $\mathbf{X_i}$ and $\mathbf{X_j}$, then $\mathbf{X_i}$ and $\mathbf{X_j}$ share the only independent component induced by the latent variable. According to Theorem \ref{th:condition}, $(\mathbf{X_i}, \mathbf{X_j})$ satisfies the One-Latent-Component condition. \qed

\subsection{Proof of Corollary \ref{pro:descendant}}

\textbf{Corollary \ref{pro:descendant}}. Suppose Assumptions \ref{asm:pure} - \ref{asm:faithful} hold. Let $X_i$ and $X_j$ be two dependent observed variables following the canonical lvLiNGAM model. If $(\{ X_i \}, \{X_i, X_j\})$ satisfies the One-Latent-Component condition, then $X_i$ is causally earlier than $X_j$.

\textit{Proof}. We will prove it by contradiction. Supposed that $X_i$ is not causally earlier than $X_j$ and $X_i$ and $X_j$ are dependent. Then $X_i$ and $X_j$ must share at least one latent component. Assume faithfulness holds, $X_i$ must be affected by latent component $S_{X_i}$. So $\{ X_i \}$ and $\{X_i, X_j\}$ share more than one latent components. According to Theorem \ref{th:condition}, $(\{ X_i \}, \{X_i, X_j\})$ won't satisfy the One-Latent-Component condition. \qed

\subsection{Proof of Lemma \ref{th:ext-one-latent-component}}

To prove Lemma \ref{th:ext-one-latent-component}, let us provide the following lemmas first.

\begin{lemma}
    Suppose Assumptions \ref{asm:pure} - \ref{asm:threechild} hold. Assume the observed variables follow the canonical lvLiNGAM model. Let $\mathbf{S_s}$ be the independent components that affect more than one observed variable. Then $\forall S_i, S_j \in \mathbf{S_s}, Aff(S_i) \neq Aff(S_j)$, where $Aff(S_i)$ is denoted as a set of observed variables affected by $S_i$.
    \label{lemma:share-diff-observed}
\end{lemma}
\textit{proof}. Assume the observed variables follow the canonical lvLiNGAM model. Let $\mathbf{S_s}$ be the independent components that affect more than one observed variable.

% We can prove $\forall S_i, S_j \in \mathbf{S_s}, Aff(S_i) \neq Aff(S_j)$, denote by $Aff(S_i)$ the observed variables affected by $S_i$. We will prove that there are no two different independent components affecting the same set of observed variables. 

The independent components contain the latent variables and the noises of observed variables, so there are three cases to consider.

\begin{enumerate}
\item Suppose both two different independent components, $S_i$ and $S_j$, are the noises of observed variables.

% According to the acyclic assumptions of data generation, all the noises of observed variables affect different observed variable sets.

According to the data generating model, each observed variable has its own noise which is independent of other noises. For any pair of observed variables, even when there is a directed edge between them, there still exists at least one noise only affects the variable that is causally earlier than the other. That is, $Aff(S_i) \neq Aff(S_j)$. 

% According to the data generation model, there is a causal order among the observed variables. For each noise of observed variables, they only affect the observed variables causal later than them. Further, each observed variable must be affected by its own noise. From the above analysis, all the noise of observed variables affects different observed variable sets.

\item Suppose both two different independent components, $S_i$ and $S_j$, are latent variables.

According to Assumption \ref{asm:pure}, we know that for each latent variable, there is at least one observed variable only affected by that latent variable. That means, there exist a $X_i$ is in $Aff(S_i)$ but not in $Aff(S_j)$. So $Aff(S_i) \neq Aff(S_j)$.

\item Suppose two different independent components consist of a latent variable $S_i$ and a noise of an observed variable $S_j$.

1) If the observed variables are influenced by the same latent variables $S_i$, according to Assumption \ref{asm:pure}, there must exist a pair of observed variables $X_i$ and $X_j$ that have no direct path between them. So the noises only influence their corresponding observed variables. That is, $S_j$ only affects $X_j$ but does not affect $X_i$. 

2) If the observed variables are influenced by the same noises $S_j$, then there must exist a direct path from $X_j$ to $X_i$. According to Assumption \ref{asm:pure}, we know that for each latent variable, there is at least one observed variable that is only affected by that latent variable. That is, there is an extra observed variable that is affected by $S_i$ but not affected by others. 
\end{enumerate}

From the above analysis, we can obtain that: $\forall S_i, S_j \in \mathbf{S_s}, Aff(S_i) \neq Aff(S_j)$, where $Aff(S_i)$.
%We can know that if there exist two observed variables $X_i, X_j$ that are affected by both the same independent components induced by both the latent variable and the observed variable, there must exist another observed variable that is not affected by the independent components induced by some observed variables. 
\qed
% 根据Assumptions \ref{asm:pure}，在观察变量受到相同隐变量影响时，一定存在一对它们之间没有有向路径的观察变量对。 我们可知如果存在两个观察变量$\mathbf{X}$同时受到相同隐变量和观察变量诱导出的独立成分$$影响时，一定存在另外一个观察变量不受部分观察变量诱导出来的独立成分的影响。
% 证毕
% 2. 
% 根据上面的分析，对于dependent的变量，一定存在一个变量与其他变量只共享

\begin{lemma}
Let $\mathbf{X_i}$ and $\mathbf{X_j}$ be the observed variable set following the canonical lvLiNGAM model, and $\mathbf{S}$ be all the independent components. We divide $\mathbf{S}$ into two disjoint subsets $\mathbf{S_1}$ and $\mathbf{S_2}$, i.e., $\mathbf{S_1} \cup \mathbf{S_2} = \mathbf{S}$ and $\mathbf{S_1} \cap \mathbf{S_2} = \emptyset$. We can rewrite the generating process of $\mathbf{X_i}$ and $\mathbf{X_j}$ as follows:
\begin{equation}
    \begin{aligned}
        \mathbf{X_i} &= \mathbf{A_{\mathbf{X_i},\mathbf{S}}} \mathbf{S} = \mathbf{A}_{\mathbf{X_i} ,\mathbf{S_1}}\mathbf{S_1} + \mathbf{A}_{\mathbf{X_i} ,\mathbf{S_2}}\mathbf{S_2},\\
        \mathbf{X_j} &= \mathbf{A_{\mathbf{X_j},\mathbf{S}}} \mathbf{S} = \mathbf{A}_{\mathbf{X_j} ,\mathbf{S_1}}\mathbf{S_1} + \mathbf{A}_{\mathbf{X_j} ,\mathbf{S_2}}\mathbf{S_2}.\\
    \end{aligned}
\end{equation}

Assume that the rows of $\mathbf{A}_{\mathbf{Y} ,\mathbf{S_1}}$ are not all zeros. $\omega^\top\mathbf{X_j} $ is independent of $\mathbf{X_i}$, if the following conditions are met: 1) $ Dim(\mathbf{X_j}) > Rank(\mathbf{A}_{\mathbf{X_j} ,\mathbf{S_1}})$; 2) $\mathbf{A}_{\mathbf{X_i}, \mathbf{S_2}} \mathbf{A}_{\mathbf{X_j},\mathbf{S_2}}^{\top} =\mathbf{0}$.
\label{th:res_condition}
\end{lemma}

\textit{Proof}. If we can find a non-zero vector $\omega$ such that $\omega^\top\mathbf{A}_{\mathbf{X_j} ,\mathbf{S_1}} =\mathbf{0}$, then
\begin{equation}
   \omega^\top\mathbf{X_j} =\omega^\top \mathbf{A}_{\mathbf{X_j} ,\mathbf{S_1}}\mathbf{S_1} + \omega^\top\mathbf{A}_{\mathbf{X_j} ,\mathbf{S_2}}\mathbf{S_2} = \omega^\top\mathbf{A}_{\mathbf{X_j} ,\mathbf{S_2}}\mathbf{S_2},
\end{equation}
which will be independent of $\mathbf{X_i}$ in light of conditions 2).

We now construct the vector $\omega$. Because of condition 1), there must exist a non-zero vector $\omega$, determined by $\mathbf{A}_{\mathbf{X_j} ,\mathbf{S_1}}$, such that $\mathbf{\omega}^{\top}\mathbf{A_{X_j, S_1}}=\mathbf{0}$, and $\omega^\top\mathbf{X_j} = \omega^\top\mathbf{A}_{\mathbf{X_j} ,\mathbf{S_2}}\mathbf{S_2}$, which is independent of $\mathbf{X_i}$. Thus, the theorem holds. \qed

\begin{lemma}
Suppose Assumptions \ref{asm:pure} - \ref{asm:faithful} hold. Let $X_i$ and $X_j$ be two observed variables following the canonical lvLiNGAM model. Let $\mathbf{S'}$ be the subset of the independent components shared by $X_i$ and $X_j$. There must exist an observed variable set $\mathbf{X_k}$ such that the influence of $\mathbf{S'}$ on $X_j$ can be removed by utilizing the surrogate regression.
\label{lemma:removable}
\end{lemma}

\textit{proof}. We need to consider two kinds of independent components, which are latent variables and the noises of observed variables.

\begin{enumerate}
    \item If $\mathbf{S'}$ contains the noises of the observed variables, each observed variable corresponding to its noise can be used as the surrogate variables of the noise, e.g., the observed variable $X_i$ is a surrogate variable of the independent component $S_{X_i}$. 
    \item If $\mathbf{S'}$ contains latent variables, based on Assumption \ref{asm:pure}, there must exist a pure set of observed variables $\mathbf{X_k}$ that can be used as surrogate variables of each latent variable.
\end{enumerate} 

%$\mathbf{S'}$ contains the independent components induced by the observed variables. There must exist the observed variables corresponding to the independent components, e.g., the observed variable $X_i$ corresponding to the independent component $S_{X_i}$. 

%Case 2) $\mathbf{S'}$ contains the independent component induced by the latent variables. Suppose Assumption \ref{asm:pure} holds, there must exist a pure observed variable $X_k'$ corresponding to the independent component.

From the above analysis, there exists an observed variable set $\mathbf{X_k}$ such that $Dim(\{X_j, \mathbf{X_k}\}) > Rank(\mathbf{A}_{\{X_j, \mathbf{X_k}\}, \mathbf{S'}})$ for any $\mathbf{S'}$. According to the Lemma \ref{th:res_condition}, there must exist a non-zero weight vector $\omega$, determined by $\mathbf{A}_{\{X_j, \mathbf{X_k}\}, \mathbf{S'}}$, such that $\mathbf{\omega}^{\top}\mathbf{A}_{\{X_j, \mathbf{X_k}\}, \mathbf{S'}} = \mathbf{0}$. Hence, there must exist an observed variable set $\mathbf{X_k}$ such that $\mathbf{S'}$ can be removed from $X_j$ by utilizing the surrogate regression. \qed

\textbf{Lemma \ref{th:ext-one-latent-component}}.
Suppose Assumptions \ref{asm:pure} - \ref{asm:faithful} hold. If there exist unidentified shared independent components, there must exist a pair of observed variable sets $\mathbf{X_i}$ and $\mathbf{X_j}$, and the surrogate variables of their identified independent components $\mathbf{S'}$ such that $(\mathbf{X_i}, \mathbf{\widetilde{X}_j})$ satisfies the One-Latent-Component condition, where $\mathbf{\widetilde{X}_j}$ are the surrogate regression residuals of $\mathbf{X_j}$ on $\mathbf{S'}$.

\textit{proof}. Let $\mathbf{X_k}$ be the surrogate variable set of $\mathbf{S'}$ in the surrogate regression. We will prove it by induction.
% 数学归纳法
% n=1

1) Assumed that there is only one independent component shared by $\mathbf{X_i}$ and $\mathbf{X_j}$. That means, $\mathbf{S}'=\emptyset$, $\mathbf{X_k}=\emptyset$, $\mathbf{\widetilde{\mathbf{X}}_j}=\mathbf{X_j}$. Suppose Assumptions \ref{asm:pure} - \ref{asm:threechild} hold. There must exist a pair of variable set $\{\mathbf{X_i}, \mathbf{X_j}\}$, and $\mathbf{X_i}$ and $\mathbf{X_j}$ follow the One-Latent-Component structure. We can identify the shared independent component according to Theorem \ref{th:condition}.

% n=2
2) Assumed that there are two independent components $\mathbf{S_2} = \{S_{21}, S_{22} \}$ shared by $\mathbf{X_i}$ and $\mathbf{X_j}$.

We can always find an observed variable set $\mathbf{X_k}$ that could remove the influence of one of the independent components from $\mathbf{X_j}$. According to Lemma \ref{lemma:share-diff-observed}, there must have an observed variable only affected by one of these two independent components $\mathbf{S_2}$. We use $\mathbf{X_{k_1}}$ to denote the observed variable set that shares one independent component (denoted by $\mathbf{S_2}$) with $\{\mathbf{X_i}, \mathbf{X_j}\}$. We can directly identify only one independent component $\mathbf{S_2} \setminus \{S_{22}\}$ shared by $\mathbf{X_{k_1}}$ and $\{\mathbf{X_i}, \mathbf{X_j}\}$ by the One-Latent-Component condition, and we can estimate the mixing matrix $\mathbf{A}_{\{\mathbf{X_i}, \mathbf{X_j}, \mathbf{X_{k_1}}\},S_{21}}$ for $\{\mathbf{X_i}, \mathbf{X_j}, \mathbf{X_{k_1}}\}$. Then their generating process can be rewritten as follows:
\begin{equation}
    \begin{aligned}
        \begin{bmatrix}
        \mathbf{X_i}\\
        \mathbf{X_j}\\
        \mathbf{X_{k_1}}\\
        \end{bmatrix}
        =
        \begin{bmatrix}
        \mathbf{A}_{\mathbf{X_i},   S_{21}} & \mathbf{A}_{\mathbf{X_i}, S_{22}}\\
        \mathbf{A}_{\mathbf{X_j},   S_{21}} & \mathbf{A}_{\mathbf{X_j}, S_{22}}\\
        \mathbf{A}_{\mathbf{X_{k_1}}, S_{21}} & \mathbf{0}\\
        \end{bmatrix}
        \begin{bmatrix}
        S_{21}\\
        S_{22}
        \end{bmatrix}
        +
        \begin{bmatrix}
        \mathbf{S'_i}\\
        \mathbf{S'_j}\\
        \mathbf{S'_{k_1}}\\
        \end{bmatrix},
    \end{aligned}
\end{equation}
where $\mathbf{S'_i}, \mathbf{S'_j}, \mathbf{S'_{k_1}}$ are independent of each others. Next, let $\widetilde{X}_j'$ be the surrogate regression residual of $X_j'$ on $S_{21}$ by utilizing $\mathbf{X_{k_1}}$ as surrogate variables. For all $X_j'$ in $\mathbf{X_j}$, we can obtain $\widetilde{X}_j'$ according to Lemma \ref{lemma:removable}. Then we have $(\mathbf{X_i}, \mathbf{\widetilde{X}_j})$ (where $\mathbf{\widetilde{X}_j} = \{\widetilde{X}_j', \dots \}$ $(\mathbf{X_i}, \mathbf{\widetilde{X}_j})$ satisfies One-Latent-Component condition, so $\mathbf{X_i}$ and $\mathbf{\widetilde{X}_j}$ follow the One-Latent-Component structure according to Theorem \ref{th:condition}. So when $\mathbf{X_i}$ and $\mathbf{X_j}$ share two independent components $\mathbf{S_2}$ and one of these two independent components $\mathbf{S_2}$ have been identified, there must exist $\mathbf{X_{k_1}}$ such that $\mathbf{X_i}$ and $\mathbf{\widetilde{X}_j}$ follow the One-Latent-Component structure.

% n=m时，n=m+1也成立
3) Assumed that there are $n$ independent components $\mathbf{S_n} = \{S_{n1}, \dots, S_{nn} \}$ shared by $\mathbf{X_i}$ and $\mathbf{X_j}$.

We can still find an observed variable set $\mathbf{X_k}$ that could remove the influence of one of the independent components from $\mathbf{X_j}$. According to Lemma \ref{lemma:share-diff-observed}, there must have an observed variable that will only be affected by $n-1$ of these $n$ independent components $\mathbf{S_n}$. We use $\mathbf{X_{k_{n-1}}}$ to denote the observed variable set that shares one independent component (denotes by $\mathbf{S_n} \setminus \{S_{nn}\} $) with $\{\mathbf{X_i}, \mathbf{X_j}\}$. We can identify $n-1$ independent components $\mathbf{S_n} \setminus \{S_{nn}\}$ shared by $\{\mathbf{X_i}, \mathbf{X_j}\}$ and $\mathbf{X_{k_{n-1}}}$ with the One-Latent-Component condition when in the case that $n-1$ independent component shared by two observed variable set, and we can estimate the mixing matrix $\mathbf{A}_{\{\mathbf{X_i}, \mathbf{X_j}, \mathbf{X_{k_{n-1}}}\}, \mathbf{S_n} \setminus \{S_{nn}\}}$ for $\{\mathbf{X_i}, \mathbf{X_j}, \mathbf{X_{k_{n-1}}}\}$. We can rewrite their generating process as follows:
\begin{equation}
    \begin{aligned}
        \begin{bmatrix}
        \mathbf{X_i}\\
        \mathbf{X_j}\\
        \mathbf{X_{k_{n-1}}}\\
        \end{bmatrix}
        =
        \begin{bmatrix}
        \mathbf{A}_{\mathbf{X_i},   \mathbf{S_n} \setminus \{S_{nn}\}} & \mathbf{A}_{\mathbf{X_i}, S_{nn}}\\
        \mathbf{A}_{\mathbf{X_j},   \mathbf{S_n} \setminus \{S_{nn}\}} & \mathbf{A}_{\mathbf{X_j}, S_{nn}}\\
        \mathbf{A}_{\mathbf{X_{k_{n-1}}}, \mathbf{S_n} \setminus \{S_{nn}\}} & \mathbf{0}\\
        \end{bmatrix}
        \begin{bmatrix}
        \mathbf{S_n} \setminus \{S_{nn}\}\\
        S_{nn}
        \end{bmatrix}
        +
        \begin{bmatrix}
        \mathbf{S'_i}\\
        \mathbf{S'_j}\\
        \mathbf{S'_{k_{n-1}}}\\
        \end{bmatrix},
    \end{aligned}
\end{equation}
where $\mathbf{S'_i}, \mathbf{S'_j}, \mathbf{S'_{k_{n-1}}}$ are independent of each others. Next, let $\widetilde{X}_j'$ be the surrogate regression residual of $X_j'$ on $\mathbf{S_n} \setminus \{S_{nn}\}$ by utilizing $\mathbf{X_{k_{n-1}}}$ as surrogate variables. For all $X_j'$ in $\mathbf{X_j}$, we can obtain $\widetilde{X}_j'$ according to Lemma \ref{lemma:removable}. Then we get $\mathbf{X_i}$ and $\mathbf{\widetilde{X}_j} = \{\widetilde{X}_j', \dots \}$ sharing only independent component $S_{22}$, so $\mathbf{X_i}$ and $\mathbf{\widetilde{X}_j}$ follow the One-Latent-Component structure. So when $\mathbf{X_i}$ and $\mathbf{X_j}$ share $n$ independent components $\mathbf{S_n}$ and $n-1$ of these $n$ independent components $\mathbf{S_n}$ have been identified, there must exist $\mathbf{X_{k_{n-1}}}$ such that $\mathbf{X_i}$ and $\mathbf{\widetilde{X}_j}$ follow the One-Latent-Component structure. 

From the above analysis, when there are unidentified independent components, there must exist a pair of variable sets $\{\mathbf{X_i}, \mathbf{X_j}\}$ and a surrogate variable set $\mathbf{X_{k}}$ such that $\mathbf{X_i}$ and $\mathbf{\widetilde{X}_j}$ follow the One-Latent-Component structure, whatever how many shared independent components have been currently identified, where $\mathbf{\widetilde{X}_j}$ are the surrogate regression residuals of $\mathbf{X_j}$ on the shared identified independent components by utilizing $\mathbf{X_{k}}$ as surrogate variables.

For the unidentified independent component in the unidentified One-Latent-Component structure, there are two cases to consider.

Case 1) The unidentified independent component is the noise of the observed variable. There must exist two observed variables influenced by this unidentified independent component, where two observed variables have a direct path between them. Hence, there must exist a pair of observed variables $\mathbf{X_i}$ and $\mathbf{X_j}$ such that $(\mathbf{X_i}, \mathbf{\widetilde{X}_j})$ satisfies the One-Latent-Component condition according to the Corollary \ref{pro:descendant}. 

Case 2) The unidentified independent component is the latent variable. Suppose Assumption \ref{asm:threechild} holds, there must exist three observed variables influenced by this unidentified independent component. Suppose Assumption \ref{asm:pure} holds, there must exist a variable set $\mathbf{\widetilde{X}_j}$ whose dimension is greater than 2. Hence, there must exist a pair of observed variables $\mathbf{X_i}$ and $\mathbf{X_j}$ such that $(\mathbf{X_i}, \mathbf{\widetilde{X}_j})$ satisfies the One-Latent-Component condition according to the Corollary \ref{pro:non-descendant}. 

From the above analysis, there must exist a pair of variable sets $\{\mathbf{X_i}, \mathbf{X_j}\}$ and the surrogate variables of $\mathbf{S'}$ such that the One-Latent-Component condition holds for $(\mathbf{X_i}, \mathbf{\widetilde{X}_j})$.\qed

\subsection{Proof of Proposition \ref{pro:purify_parent}}

% Before presenting Proposition \ref{pro:purify_parent}, let us introduce the following lemmas, which will be needed in the proof of Proposition \ref{pro:purify_parent}.

% \begin{lemma}
% Let $X_i$ and $U$ be two variables in LiNGAM model. We can rewrite the mixing coefficient $\mathbf{A}_{W, S_U}$ between the observed variable (mixture) $W$ and the independent component $S_U$ related to the variable $U$ can be rewritten as:
% \begin{equation}
%     \begin{aligned}
%         \mathbf{A}_{W, S_U} = \sum_{P \in \mathcal{P}(U, W)} \prod_{l \rightarrow k \in P}\mathbf{B}_{k,l},
%     \end{aligned}
%     % \label{eq:proof_3_5_2}
% \end{equation}
% where $\mathcal{P}(U, W)$ denotes the set of directed paths from $U$ to $W$, $l \rightarrow k$ denotes as a directed edge from $l$ to $k$ and $\mathbf{B}_{k,l}$ denotes the causal coefficient of $l$ on $k$.
% \label{lemma:causal_path}
% \end{lemma}

\textbf{Proposition \ref{pro:purify_parent}}. Suppose Assumptions \ref{asm:pure} - \ref{asm:faithful} hold. Let $X_i$ and $X_j$ be two observed variables following the canonical lvLiNGAM model and satisfying $X_i$ is causally earlier than $X_j$. There is no directed edge between $X_i$ and $X_j$, if and only if there exists an observed variable set $\mathbf{X_k}$ satisfying that 1)$X_i, X_j \notin \mathbf{X_k}$; 2)$\forall X_k' \in \mathbf{X_k}$, if $X_k'$ is a descendant of $X_i$, then $X_k'$ must be an ancestor of $X_j$; 3) $\widetilde{X}_j'$ and $\widetilde{X}_j''$ are independent of $X_i$, where $\widetilde{X}_j'$ and $\widetilde{X}_j''$ are the surrogate regression residuals of $X_j$ on the independent components by utilizing $\mathbf{X_k}$ and $\{X_i,\mathbf{X_k} \}$ as surrogates, respectively.

\textit{Proof}. 
% According to the Lemma \ref{lemma:causal_path}, 
Let $X_i$ and $X_j$ be two variables following the lvLiNGAM model and $\mathbf{B}_{k,l}$ be the causal coefficient of $l$ on $k$. Let $\mathcal{P}(X_j, X_i)$ denotes the set of directed paths from $X_j$ to $X_i$, $X_l \rightarrow X_k$ denotes as a directed edge from $X_l$ to $X_k$. The mixing coefficient $\mathbf{A}_{X_j, S_{X_i}}$ between $X_j$ and the independent component $S_{X_i}$ related to the $X_i$ can be rewritten as:
\begin{equation}
    \begin{aligned}
        \mathbf{A}_{X_j, S_{X_i}} &= \sum_{P \in \mathcal{P}(X_i, X_j)} \prod_{X_l \rightarrow X_k \in P}\mathbf{B}_{X_k, X_l}\\
        &= \sum_{P \in \mathcal{P}(X_i, X_j)\setminus <X_i,X_j>} \prod_{X_l \rightarrow X_k \in P}\mathbf{B}_{X_k, X_l} + \mathbf{B}_{X_j, X_i},
    \end{aligned}
\end{equation}
where $<X_i, X_j>$ is denotes as a directed path from $X_i$ to $X_j$ that only contains two variables $X_i$ and $X_j$. Let $Pa(X_j)$ be the parents of $X_j$. Then, $\forall P_{X_j} \in Pa(X_j),~ \mathbf{A}_{P_{X_j}, S_{X_i}}=\sum_{P \in \mathcal{P}(X_i, P_{X_j})} \prod_{l \rightarrow k \in P}\mathbf{B}_{k,l}$ and $\mathbf{A}_{X_j, S_{X_i}} = \sum_{P_{X_j} \in Pa(X_j)} \mathbf{B}_{X_j, P_{X_i}} \mathbf{A}_{P_{X_j}, S_{X_i}}$. 

% So we can rewrite the mixing coefficient $\mathbf{A}_{X_j, S_{X_i}}$ as follows:
% \begin{equation}
%     \begin{aligned}
%         \mathbf{A}_{X_j, S_{X_i}} &= \sum_{P \in \mathcal{P}(X_i, X_j)} \prod_{l \rightarrow k \in P}\mathbf{B}_{k,l}\\
%             &= \sum_{P \in \mathcal{P}(X_i, X_j)\setminus <X_i,X_j>} \prod_{l \rightarrow k \in P}\mathbf{B}_{k, l} + \mathbf{B}_{X_j, X_i},
%     \end{aligned}
% \end{equation}
% where $<X_i, X_j>$ denotes as a directed path from $X_i$ to $X_j$ which only contains $\{X_i, X_j\}$. Let $Pa(X_j)$ be the parents of $X_j$, there are $\forall P_{X_j} \in Pa(X_j),~ \mathbf{A}_{P_{X_j}, S_{X_i}}=\sum_{P \in \mathcal{P}(X_i, P_{X_j})} \prod_{l \rightarrow k \in P}\mathbf{B}_{k,l}$ and $\mathbf{A}_{X_j, S_{X_i}} = \sum_{P_{X_j} \in Pa(X_j)} \mathbf{B}_{X_j, P_{X_i}} \mathbf{A}_{P_{X_j}, S_{X_i}}$. 

Let $\mathbf{S_i}$ be the independent components of $X_i$. There are two cases to consider.

Case 1) Assume that there is a directed edge from $X_i$ to $X_j$, then $\mathbf{B}_{X_j, X_i} \neq 0$. Because of condition 1) and 2), if $\mathbf{A}_{X_k', S_{X_i}} \neq 0$, then $\mathbf{A}_{X_j, S_{X_k'}} \neq 0$. Furthermore, $\mathbf{A}_{X_j, S_{X_i}} = \sum_{P \in \mathcal{P}(X_i, X_j)\setminus <X_i, X_j>} \prod_{l \rightarrow k \in P}\mathbf{B}_{k, l} + \mathbf{B}_{X_j, X_i}$ and $\mathbf{A}_{X_j, \mathbf{S_i}}$ is not a linear combination of $\mathbf{A}_{\mathbf{X_k}, \mathbf{S_i}}$. Hence, the vectors in $\mathbf{A}_{\{X_j, \mathbf{X_k}\}, \mathbf{S_i}}$ are linearly dependent, there does not exist $\omega \neq \mathbf{0}$ such that $\omega^\top\mathbf{A}_{\{X_j, \mathbf{X_k}\}, \mathbf{S_i}}=\mathbf{0}$, which violate the condition in Lemma \ref{th:res_condition}. From the above analysis, $\widetilde{X}_i'$ wouldn't be independent of $X_i$.

Case 2) Assume that there is no directed edge from $X_i$ to $X_j$, then $\mathbf{B}_{X_j, X_i} = \mathbf{0}$. Because of condition 1) and 2), if $\mathbf{A}_{X_k', S_{X_i}} \neq 0$, then $\mathbf{A}_{X_j, S_{X_k'}} \neq 0$. Furthermore, we have $\mathbf{A}_{X_j, S_{X_i}} = \sum_{P \in \mathcal{P}(X_i , X_j)\setminus <X_i, X_j>} \prod_{l \rightarrow k \in P}\mathbf{B}_{k, l}$, and $\mathbf{A}_{X_j, \mathbf{S_i}}$ is a linear combination of $\mathbf{A}_{\mathbf{X_k}, \mathbf{S_i}}$. So there must exist an observed variable set $\mathbf{X_k}$ satisfying $\mathbf{A}_{X_j, S_{X_i}} = \mathbf{B}_{X_j, \mathbf{X_k}} \mathbf{A}_{\mathbf{X_k}, S_{X_i}}$, i.e., $\mathbf{A}_{X_j, S_{X_i}}$ is a linear combination of $\mathbf{A}_{\mathbf{X_k}, S_{X_i}}$. Further, $\mathbf{A}_{X_j, \mathbf{S_i}}$ is a linear combination of $\mathbf{A}_{\mathbf{X_k}, \mathbf{S_i}}$. Because $X_i$ is causally earlier than $X_j$, then $Dim(\{X_i, X_j, \mathbf{X_k}\}) > Rank(\mathbf{A}_{\{X_i, X_j, \mathbf{X_k}\}, \mathbf{S_i}})$ according to Lemma \ref{lemma:removable}. It also holds that $Dim(\{X_j, \mathbf{X_k}\}) > Rank(\mathbf{A}_{\{X_j, \mathbf{X_k}\}, \mathbf{S_i}})$. Then there exist $\omega$ such that $\omega^\top\mathbf{A}_{\{X_j, \mathbf{X_k}\}, \mathbf{S_i}} = \mathbf{0}$ and $\omega \neq \mathbf{0}$. 
Thus, $\mathbf{X_k}$ can be used as surrogate variables of $\mathbf{S_i}$ to remove the influence of $\mathbf{S_i}$ on $X_j$, so the corresponding surrogate regression residual $\widetilde{X}_j'$ is independent of $X_i$. Similarly, by using $\{X_j,\mathbf{X_k}\}$ as surrogate variables, we can obtain the surrogate regression residual $\widetilde{X}_j''$ that does not contain the influence of $\mathbf{S_i}$, which makes the $\widetilde{X}_j''$ and $X_i$ independent.
%So we can utilize simply $X_j$ and $\mathbf{X_k}$ to remove all the independent components that affect the variable $\mathbf{S_i}$ according to the Lemma \ref{th:res_condition}. From the above analysis, $\widetilde{X}_j'$ and $\widetilde{X}_j''$ would be independent of $X_i$. \qed

% If the condition 2) holds, then $Dim(\{X_i, X_j, \mathbf{X_k}\}) = Rank(\mathbf{A}_{\{X_i, X_j, \mathbf{X_k}\}, \mathbf{S_i}}) + 2 = Rank(\mathbf{A}_{\{X_j, \mathbf{X_k}\}, \mathbf{S_i}}) + 2$. Hence, the condition 1) of \ref{th:res_condition}, i.e., $Dim(\{X_j, \mathbf{X_k}\}) > Rank(\mathbf{A}_{\{X_j, \mathbf{X_k}\}, \mathbf{S_i}})$ holds. Because of $\mathbf{S_i}$ is all the independent components that affect the variable $X_i$, the condition 2) of \ref{th:res_condition} holds. So we can utilize simply $X_j$ and $\mathbf{X_k}$ to remove all the independent components that affect the variable $\mathbf{S_i}$. From the above analysis, $\widetilde{X}_j'$ and $\widetilde{X}_j''$ would independent of $X_i$. \qed

\subsection{Proof of Theorem \ref{th:identify}}

\textbf{Theorem \ref{th:identify}}. Suppose Assumptions \ref{asm:pure} - \ref{asm:faithful} hold, and that the data is generated by the canonical lvLiNGAM model. Then the causal structure as well as the causal coefficients can be identified by Algorithm \ref{alg:algorithm_framework}.

\textit{Proof.} Supposed that Assumptions \ref{asm:pure} - \ref{asm:faithful} hold, and the data is generated by the canonical lvLiNGAM model. According to Theorem \ref{th:condition} and Lemma \ref{th:ext-one-latent-component}, we can utilize the One-Latent-Component condition to identify all shared independent components. Thus, the causal coefficient can be identified. According to Corollary \ref{pro:non-descendant}, we can identify how the latent variables affect the observed variables. According to Corollary \ref{pro:descendant}, we can identify the causal order between the observed variables. According to Proposition \ref{pro:purify_parent}, we can remove the redundant edges from causal order. Thus, the causal structure is identified. \qed

\section{Closed-Form Solution to OICA in Specific Cases by General Higher-order Cumulants}\label{app:est_by_general_cum}

Note that we have to rely on the second-order cumulant (equaling covariance) which is inevitable. Specifically, on the Gaussian assumption, we can only use second-order information. One makes use of higher-order statistics. First of all, it is worth mentioning that our result can be applied to each of the higher-order cumulants. Using other orders of cumulants will not hold further technical challenges. However, generally speaking, the higher order of the cumulants is more sensitive to outliers. Furthermore, a lot of distributions may be symmetric with the zero third-order cumulants, while their non-Gaussianity can be reflected by the four-order cumulant. Besides, fourth-order cumulants would not be zero for most non-Gaussian distributions \citep{hyvarinen2001independent}. So in this paper, we decide to use the fourth-order cumulant. If needed, in a specific case, one may use cumulants of other orders or even their combinations as well.

We further provide a detailed procedure to use other typical higher-order cumulants as follows. For convenience, let $\mathcal{C}_{a}(X_i) = cum(\underbrace{X_i,\dots}_{a \,\text{times}})$, $\mathcal{C}_{a,b}(X_i, X_j) = cum(\underbrace{X_i,\dots}_{a \,\text{times}}, \underbrace{X_j,\dots}_{b  \,\text{times}})$, where $a > 0$ and $b > 0$. For example, $\mathcal{C}_{3}(X_i) = cum(X_i, X_i, X_i)$,  $\mathcal{C}_{2,1}(X_i, X_j) = cum(X_i, X_i, X_j)$. The following theorem can be seen as the generalization of Theorem \ref{th:shared-one-exogenous}.

\begin{theorem} Let $X_i$ and $X_j$ be two observed variables following Eq. (\ref{eq:ica}). Suppose $X_i$ and $X_j$ follow the One-Latent-Component structure and that $S$ is the only one shared non-Gaussian latent component of them and has a unit variance. Further suppose there exists $n$ and $m$ such that $\mathcal{C}_{n, m + 1}(X_i, X_j)\neq 0 $ and $\mathcal{C}_{n + 1, m}(X_i, X_j) \neq 0$ where $n > 0$ and $m > 0$. The mixing coefficients between $\{X_i, X_j\}$ and $S$, denoted by $\hat{\alpha}_i$ and $\hat{\alpha}_j$ respectively, can be identified by the higher-order cumulant as follows:
\begin{equation}
\begin{aligned}
\hat{\alpha}_i &= \sqrt{\frac{\mathcal{C}_{n, m + 1}(X_i, X_j)}{\mathcal{C}_{n + 1, m}(X_i, X_j)}\mathcal{C}_{1, 1}(X_i, X_j)},\\
\hat{\alpha}_j &= \frac{\mathcal{C}_{1, 1}(X_i, X_j)}{\hat{\alpha}_i}.
\end{aligned}
\label{eq:general_mixing_coef}
\end{equation}
\label{th:general_estimation}
\end{theorem}

\textit{Proof}. Similar to the proof of \ref{th:shared-one-exogenous}, we consider the cumulant \( (X_i, X_j)\), \((\underbrace{X_i,\dots}_{n \,\text{times}}, \underbrace{X_j,\dots}_{m+1  \,\text{times}})\) and \((\underbrace{X_i,\dots}_{n+1 \,\text{times}}, \underbrace{X_j,\dots}_{m  \,\text{times}})\) as:
\begin{equation}
\begin{aligned}
    \mathcal{C}_{1, 1}(X_i, X_j) &= \alpha_i\alpha_j \mathcal{C}_{2}(S), \\
    \mathcal{C}_{n, m + 1}(X_i, X_j) &= \alpha_i^{n}\alpha_j^{m + 1} \mathcal{C}_{n + m + 1}(S), \\
    \mathcal{C}_{n + 1, m}(X_i, X_j) &= \alpha_i^{n+1}\alpha_j^{m} \mathcal{C}_{n + m + 1}(S). \\
\end{aligned}
\end{equation}

Then the square of $\alpha_i$ is obtained in the following way: 
\begin{equation}
\begin{aligned}
&\frac{\mathcal{C}_{n, m + 1}(X_i, X_j)}{\mathcal{C}_{n + 1, m}(X_i, X_j)}\mathcal{C}_{1, 1}(X_i, X_j)\\
=& \frac{\alpha_i}{\alpha_j}(\alpha_i\alpha_j\sigma_{S}^2)\\
=& \alpha_i^2 \sigma_{S}^2.
\end{aligned}
\end{equation}

Thus, $\alpha_i$ and $\alpha_j$ can be estimated by the cumulants according to Eq. (\ref{eq:general_mixing_coef}), which can be easily proved according to the proof of Theorem \ref{th:shared-one-exogenous}. \qed

\section{Running Example of Our Method} \label{app:toy}

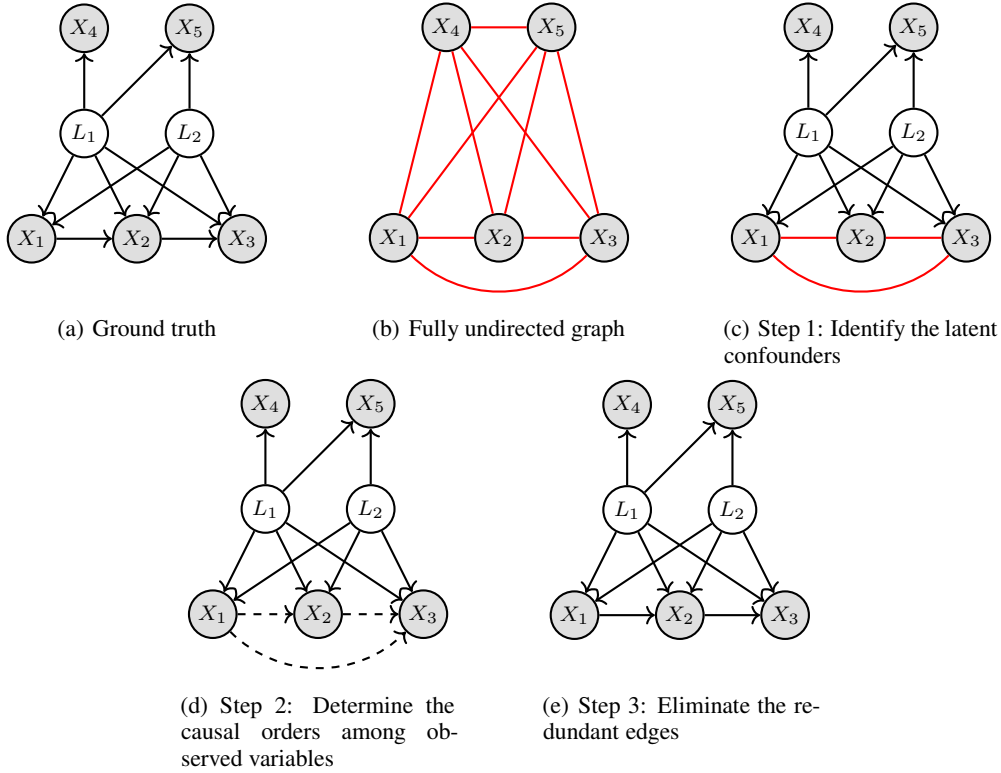
\begin{figure}[ht]
    \centering
    \subfigure[Ground truth]{
        \begin{tikzpicture}[thick]
            \node[causallatent, xshift= -0.70cm, yshift=  0.70cm] (L1) {$L_1$};
            \node[causallatent, xshift=  0.70cm, yshift=  0.70cm] (L2) {$L_2$};
            \node[causalobs,    xshift= -1.40cm, yshift= -0.70cm] (X1) {$X_1$} ;
            \node[causalobs,    xshift=  0.0cm, yshift= -0.70cm] (X2) {$X_2$} ;
            \node[causalobs,    xshift=  1.40cm, yshift= -0.70cm] (X3) {$X_3$} ;
            \node[causalobs,    xshift= -0.70cm, yshift=  2.10cm] (X4) {$X_4$} ;
            \node[causalobs,    xshift=  0.70cm, yshift=  2.10cm] (X5) {$X_5$} ;
            
            \edge {L1} {X1, X2, X3, X4, X5};
            \edge {L2} {X1, X2, X3, X5};
            \edge {X1} {X2};
            \edge {X2} {X3};
            \path (X1) edge [draw=none, bend right=45] (X3) ;
        \end{tikzpicture}
    }
    \hspace{1.0cm}
    \subfigure[Fully undirected graph]{
        \begin{tikzpicture}[thick]
            \node[causalobs,    xshift= -1.40cm, yshift= -0.7cm] (X1) {$X_1$} ;
            \node[causalobs,    xshift=  0.0cm, yshift= -0.7cm] (X2) {$X_2$} ;
            \node[causalobs,    xshift=  1.40cm, yshift= -0.7cm] (X3) {$X_3$} ;
            \node[causalobs,    xshift= -0.7cm, yshift=  2.10cm] (X4) {$X_4$} ;
            \node[causalobs,    xshift=  0.7cm, yshift=  2.10cm] (X5) {$X_5$} ;
            \uedge {X1} {X2, X4, X5};
            \uedge {X2} {X3, X4, X5};
            \uedge {X3} {X4, X5};
            \uedge {X4} {X5};
            \path (X1) edge [-, red, bend right=45] (X3) ;
        \end{tikzpicture}
    }
    \hspace{1.0cm}
    \subfigure[Step 1: Identify the latent confounders]{
        \begin{tikzpicture}[thick]
            \node[causallatent, xshift= -0.7cm, yshift=  0.7cm] (L1) {$L_1$};
            \node[causallatent, xshift=  0.7cm, yshift=  0.7cm] (L2) {$L_2$};
            \node[causalobs,    xshift= -1.40cm, yshift= -0.7cm] (X1) {$X_1$} ;
            \node[causalobs,    xshift=  0.0cm, yshift= -0.7cm] (X2) {$X_2$} ;
            \node[causalobs,    xshift=  1.40cm, yshift= -0.7cm] (X3) {$X_3$} ;
            \node[causalobs,    xshift= -0.7cm, yshift=  2.10cm] (X4) {$X_4$} ;
            \node[causalobs,    xshift=  0.7cm, yshift=  2.10cm] (X5) {$X_5$} ;
            \uedge {X1} {X2};
            \uedge {X2} {X3};
            \path (X1) edge [-, red, bend right=45] (X3) ;
            \edge {L1} {X1, X2, X3, X4, X5};
            \edge {L2} {X1, X2, X3, X5};
        \end{tikzpicture}
    }
    \hspace{1.0cm}
    \subfigure[Step 2: Determine the causal orders among observed variables]{
        \begin{tikzpicture}[thick]
            \node[causallatent, xshift= -0.7cm, yshift=  0.7cm] (L1) {$L_1$};
            \node[causallatent, xshift=  0.7cm, yshift=  0.7cm] (L2) {$L_2$};
            \node[causalobs,    xshift= -1.40cm, yshift= -0.7cm] (X1) {$X_1$} ;
            \node[causalobs,    xshift=  0.0cm, yshift= -0.7cm] (X2) {$X_2$} ;
            \node[causalobs,    xshift=  1.40cm, yshift= -0.7cm] (X3) {$X_3$} ;
            \node[causalobs,    xshift= -0.7cm, yshift=  2.10cm] (X4) {$X_4$} ;
            \node[causalobs,    xshift=  0.7cm, yshift=  2.10cm] (X5) {$X_5$} ;
            \path (X1) edge [-To, dashed] (X2) ;
            \path (X2) edge [-To, dashed] (X3) ;
            \path (X1) edge [-To, dashed, bend right=45] (X3) ;
            % \edge {X1} {X2, X3};
            % \edge {X2} {X3};
            \edge {L1} {X1, X2, X3, X4, X5};
            \edge {L2} {X1, X2, X3, X5};
        \end{tikzpicture}
    }
    \hspace{1.0cm}
    \subfigure[Step 3: Eliminate the redundant edges]{
        \begin{tikzpicture}[thick]
            \node[causallatent, xshift= -0.7cm, yshift=  0.7cm] (L1) {$L_1$};
            \node[causallatent, xshift=  0.7cm, yshift=  0.7cm] (L2) {$L_2$};
            \node[causalobs,    xshift= -1.40cm, yshift= -0.7cm] (X1) {$X_1$} ;
            \node[causalobs,    xshift=  0.0cm, yshift= -0.7cm] (X2) {$X_2$} ;
            \node[causalobs,    xshift=  1.40cm, yshift= -0.7cm] (X3) {$X_3$} ;
            \node[causalobs,    xshift= -0.7cm, yshift=  2.10cm] (X4) {$X_4$} ;
            \node[causalobs,    xshift=  0.7cm, yshift=  2.10cm] (X5) {$X_5$} ;
            \edge {X1} {X2};
            \edge {X2} {X3};
            \edge {L1} {X1, X2, X3, X4, X5};
            \edge {L2} {X1, X2, X3, X5};
            \path (X1) edge [draw=none, bend right=45] (X3) ;
        \end{tikzpicture}
    }
    \caption{A running example of the proposed algorithm on five observed variables $\{X_1, X_2, \dots, X_5\}$ and two latent confounders $\{L_1, L_2\}$.}
    \label{fig:toy}
\end{figure}

In this section, we will show how our proposed algorithm works by an example whose ground truth is shown in Figure \ref{fig:toy}(a). We start from the complete undirected graph as Figure \ref{fig:toy}(b). 

\textbf{Step 1: Identifying the latent confounders}

According to Algorithm \ref{alg:algorithm_framework}, we first randomly select two observed variables, e.g. $X_1$ and $X_2$. Second, we first randomly select an observed variable different from the previously selected observed variables, such as $X_3$. Then we assume $\{X_1, X_2, X_3\}$ are influenced by latent variable $L_1$. Third, let $\mathbf{X_i} = \{X_3\}$ and $\mathbf{X_j} = \{X_1, X_2\}$, and we use $\mathbf{X_i}$ to estimate the mixing matrix between $L_1$ and $\{X_1, X_2, X_3\}$. Based on the estimated mixing matrix, we cannot find a non-zero $\omega$ such that $\omega^{\top}\mathbf{X_j} \ci \mathbf{X_i}$, i.e., $(\{X_3\}, \{X_1, X_2\})$ violates One-Latent-Component condition. Then we move to the next iteration. After many iterations, we may find $(\{X_4\}, \{X_1, X_2\}$ satisfies the One-Latent-Component condition, then we accept the hypothetical structure, introduce a new latent confounder $L_1$ affecting $X_1$, $X_2$ and $X_3$, and remove the edges between $(\{X_1, X_2\}$ and $\{X_3\}$. 
Similarly, we can also identify $L_1$ affecting $\{X_4, X_5\}$. 

After that, we can remove the influence of $L_1$ from $X_1$ and $X_2$ by utilizing surrogate regression with $X_4$, and the surrogate regression residuals are denoted by $\widetilde{X}_1$ and $\widetilde{X}_2$ respectively. Then $\widetilde{X}_1$ and $\widetilde{X}_2$ can be used to estimate the mixing matrix between another latent component $L_2$ and $\{X_1, X_2, X_5\}$. With the situation that $(\{X_5\}, \{\widetilde{X}_1, \widetilde{X}_2\})$ satisfies the One-Latent-Component condition, we can identify the latent confounder $L_2$ affecting $\{X_1, X_2, X_5\}$. Finally, we can obtain the causal graph shown in Figure \ref{fig:toy}(c).

\textbf{Step 2: Determining the causal orders among observed variables}

Based on Step 1, we have identified the latent variables $L_1$ and $L_2$, and obtain pure sets of observed variables $\{X_1,X_2,X_3\}$, $\{X_4\}$ and $\{X_5\}$. Next, we focus on determining the causal order among $\{X_1,X_2,X_3\}$, and use the additional variables as the surrogates of latent components. 

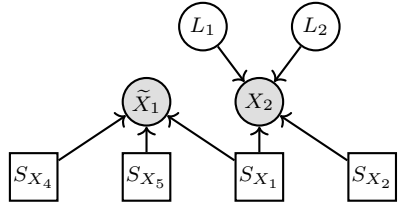
\begin{figure}
    \centering
    \begin{tikzpicture}[thick]
        \node[causallatent, xshift=  0.00cm, yshift=  1.00cm] (L1) {$L_1$};
        \node[causallatent, xshift=  1.50cm, yshift=   1.00cm] (L2) {$L_2$};
        \node[causalobs,    xshift= -0.75cm, yshift=  0.00cm] (X1) {$\widetilde{X}_1$} ;
        \node[causalobs,    xshift=  0.75cm, yshift=  0.00cm] (X2) {$X_2$} ;
        \node[causaleps,    xshift=  0.75cm, yshift= -1.00cm] (EX1) {$S_{X_1}$} ;
        \node[causaleps,    xshift=  2.25cm, yshift= -1.00cm] (EX2) {$S_{X_2}$} ;
        \node[causaleps,    xshift= -2.25cm, yshift= -1.00cm] (EX4) {$S_{X_4}$} ;
        \node[causaleps,    xshift= -0.75cm, yshift= -1.00cm] (EX5) {$S_{X_5}$} ;
        \edge {L1, L2} {X2};
        \edge {EX1} {X1, X2};
        \edge {EX2} {X2};
        \edge {EX4, EX5} {X1};
    \end{tikzpicture}
    \caption{The local causal structure after removing the influence of $L_1$ and $L_2$ on $X_1$ in Figure \ref{fig:toy}(a).}
    \label{fig:causal_order}
\end{figure}

The procedure is performed as follows. First, we assume $X_1$ is causally earlier than $X_2$. Second, we remove the influence of $\{L_1, L_2\}$ from $X_1$ and $X_2$ by utilizing surrogate regression with $\{X_4, X_5\}$, and the surrogate regression residuals are denoted by $\widetilde{X}_1$ and $\widetilde{X}_2$ respectively. The causal graph over $\widetilde{X}_1$ and ${X}_2$ is illustrated in Figure \ref{fig:causal_order}. Third, let $\mathbf{X_i} = \{X_1\}$ and $\mathbf{X_j} = \{\widetilde{X}_1, \widetilde{X}_2\}$. $\mathbf{X_i}$ is used to estimate the mixing matrix between $S_{X_1}$ and $\{X_1, X_2\}$ where $S_{X_1}$ is the noise of $X_1$. With the situation that $(\{X_1\}, \{\widetilde{X}_1, \widetilde{X}_2\})$ satisfies the One-Latent-Component condition, we can accept the hypothetical causal order that $X_1$ is causally earlier than $X_2$. Similarly, we can also determine the causal order among $\{X_1, X_2, X_3\}$. At last, we can obtain the causal graph shown in Figure \ref{fig:toy}(d).

\textbf{Step 3: Eliminating the redundant edges}

In the previous steps, we obtain the causal order among $\{X_1, X_2, X_3\}$. Naturally, one may be interested in whether there is a direct edge between two variables. If not, then the edge is redundant and can be removed. Considering $X_1$ and $X_3$, the procedure is performed as follows. We have known that both $X_1$ and $X_2$ are causally earlier than $X_3$ and $X_1$ is causally earlier than $X_2$, so we assume there is no directed edge between $X_1$ and $X_3$. According to the result of Step 2, we can remove the influence of $\{L_1, L_2, S_{X_1}\}$ from $X_3$ by utilizing surrogate regression with $\{X_2, X_4, X_5\}$ and $\{X_1, X_2, X_4, X_5\}$, and the surrogate regression residuals are denoted by $\widetilde{X}_3'$ and $\widetilde{X}_3''$ respectively. With the situation that $\widetilde{X}_3' \ci X_1$ and $\widetilde{X}_3'' \ci X_1$, we obtain that there is no directed edge between $X_1$ and $X_3$, that is, the edge between $X_1$ and $X_3$ is redundant and can be eliminated. For other pairs $\{X_1, X_3\}$ and $\{X_2, X_3\}$, we perform the similar produce described above, and cannot find the satisfactory situation provided in Proposition \ref{pro:purify_parent}. Thus, the result of this step is refined in Figure \ref{fig:toy}(e).

\section{Analysis of Estimation Variance of Fourth-order Cumulant}\label{app:cumest} 

In this paper, we use the unbiased estimator \citep{schefczik2019ready} to calculate the cumulant. Besides, the variance of estimated fourth order cumulant $C_4$ is
$$
Var(C_4) = (\mu_{8} - 12 \mu_6\mu_2 - 8 \mu_5\mu_3 - \mu_4^{2} + 48 \mu_{4}\mu_2^{2} + 64\mu_3^2-\mu_2 - 36\mu_2^4)/n,
$$
where $\mu_k$ is $k^{th}$ order central moment and $n$ is the sample size, which is provided in \citep{krasilnikov2019analysis, pandav2019effect}. From this point of view, the cumulant estimators are consistent.

\section{Experiments}\label{app:exp}

\subsection{Synthetic Data in Six Specific Cases}\label{app:exp_six_cases}

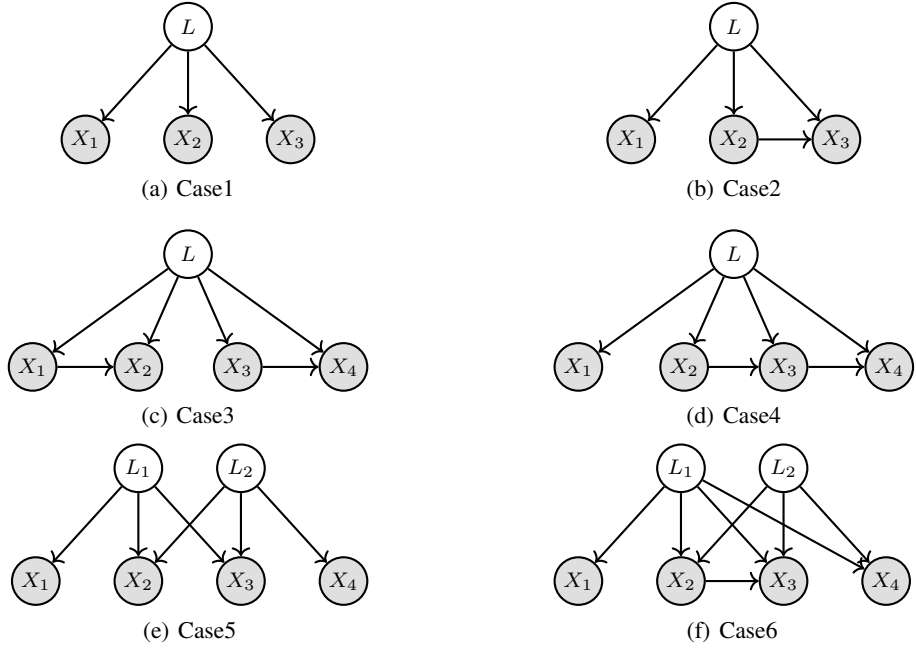
\begin{figure}[ht]
    \centering
    \subfigure[Case1]{
        \begin{tikzpicture}[thick]
            \node[causallatent] (L) {$L$}; %
            \node[causalobs, left=  20pt of L, yshift=-1.5cm] (X1) {$X_1$} ;
            \node[causalobs,                     yshift=-1.5cm] (X2) {$X_2$} ;
            \node[causalobs, right= 20pt of L, yshift=-1.5cm] (X3) {$X_3$} ;
            \edge {L} {X1, X2, X3};
        \end{tikzpicture}
    }
    \hspace{100pt}
    \subfigure[Case2]{
        \begin{tikzpicture}[thick]
            \node[causallatent] (L) {$L$}; %
            \node[causalobs, left=  20pt of L, yshift=-1.5cm] (X1) {$X_1$} ;
            \node[causalobs,                     yshift=-1.5cm] (X2) {$X_2$} ;
            \node[causalobs, right= 20pt of L, yshift=-1.5cm] (X3) {$X_3$} ;
            \edge {L} {X1, X2, X3};
            \edge {X2} {X3};
        \end{tikzpicture}
    }

    \subfigure[Case3]{
        \begin{tikzpicture}[thick]
            \node[causallatent] (L) {$L$}; %
            \node[causalobs, left=  40pt of L, yshift=-1.5cm] (X1) {$X_1$} ;
            \node[causalobs, left= 0pt of L, yshift=-1.5cm] (X2) {$X_2$} ;
            \node[causalobs, right= 0pt of L, yshift=-1.5cm] (X3) {$X_3$} ;
            \node[causalobs, right= 40pt of L, yshift=-1.5cm] (X4) {$X_4$} ;
            \edge {L} {X1, X2, X3, X4};
            \edge {X1} {X2};
            \edge {X3} {X4};
        \end{tikzpicture}
    }
    \hspace{60pt}
    \subfigure[Case4]{
        \begin{tikzpicture}[thick]
            \node[causallatent] (L) {$L$}; %
            \node[causalobs, left= 40pt of L, yshift=-1.5cm] (X1) {$X_1$} ;
            \node[causalobs, left= 0pt of L, yshift=-1.5cm] (X2) {$X_2$} ;
            \node[causalobs, right= 0pt of L, yshift=-1.5cm] (X3) {$X_3$} ;
            \node[causalobs, right= 40pt of L, yshift=-1.5cm] (X4) {$X_4$} ;
            \edge {L} {X1, X2, X3, X4};
            \edge {X2} {X3};
            \edge {X3} {X4};
        \end{tikzpicture}
    }
    \subfigure[Case5]{
        \begin{tikzpicture}[thick]
            \node[causallatent] (L1) {$L_1$}; %
            \node[causallatent, right= 20pt of L1] (L2) {$L_2$}; %
            \node[causalobs, left= 20pt of L1, yshift=-1.5cm] (X1) {$X_1$} ;
            \node[causalobs, right= 20pt of X1] (X2) {$X_2$} ;
            \node[causalobs, right= 20pt of L2, yshift=-1.5cm] (X4) {$X_4$} ;
            \node[causalobs, left= 20pt of X4] (X3) {$X_3$} ;
            \edge {L1} {X1, X2, X3};
            \edge {L2} {X2, X3, X4};
        \end{tikzpicture}
    }
    \hspace{60pt}
    \subfigure[Case6]{
        \begin{tikzpicture}[thick]
            \node[causallatent] (L1) {$L_1$}; %
            \node[causallatent, right= 20pt of L1] (L2) {$L_2$}; %
            \node[causalobs, left= 20pt of L1, yshift=-1.5cm] (X1) {$X_1$} ;
            \node[causalobs, right= 20pt of X1] (X2) {$X_2$} ;
            \node[causalobs, right= 20pt of L2, yshift=-1.5cm] (X4) {$X_4$} ;
            \node[causalobs, left= 20pt of X4] (X3) {$X_3$} ;
            \edge {L1} {X1, X2, X3, X4};
            \edge {L2} {X2, X3, X4};
            \edge {X2} {X3};
        \end{tikzpicture}
    }
    \caption{The causal graphs corresponding to six cases in the experiments on synthetic data}
    \label{fig:exp_graphs}
\end{figure}

In the simulation studies, to mimic real, complex situations, we exploited six rather complex simulation settings. In these settings, the observed data is generated according to the latent variable LiNGAM model, where the causal coefficient $b_{ij}$ is sampled from a uniform distribution between $[0.2, 0.8]$, and the noises are generated from normal distribution $(0.0,1.0)$ variables to the third power. For each model, the sample size $N$ is among $[500, 1000, 2000]$. The true causal structures are given in Figure \ref{fig:exp_graphs}. 

We compare our method with FCI \citep{spirtes2000causation}, lvLiNGAM \citep{hoyer2008estimation}, DLiNGAM \citep{shimizu2011directlingam}, PLiNGAM \citep{tashiro2014parcelingam} and RCD \citep{maeda2020rcd}. We evaluate the results in terms of directed causal edges between variables and non-adjacent relationships, by using the precision, recall and $F_1$-score as evaluation metrics. In detail, Precision is the percentage of correct directed causal/non-adjacent relationships between observed variables among all directed causal/non-adjacent relationships returned by the algorithm. Recall is the percentage of correct directed causal/non-adjacent relationships that are found by the search among true directed causal/non-adjacent relationships between observed variables. $F_1$-score is defined as $F_1 = \frac{2\times Precision \times Recall}{Precision + Recall}$. 

Besides, we use Root Mean Square Errors (RMSE) as evaluation metrics to evaluate the performance of estimating the causal coefficients and compute the average computation time to show the efficiency of our method. RMSE is estimated as
$\sqrt{\frac{1}{n^2}\sum_{i=1}^{n}\sum_{j = 1}^{n} (b_{ij}-\hat{b}_{ij})^2}$, where $b_{ij}$ is the true causal coefficient between $x_i$ and $x_j$, and $\hat{b}_{ij}$ is the estimated causal coefficient between $x_i$ and $x_j$. Each experiment was repeated 10 times with randomly generated data and the results were averaged. 

The results of different methods are illustrated in Table \ref{tab:p_causal_arrow} - Table \ref{tab:rmse}.

\textbf{Evaluation on directed causal edges.} As shown in Table \ref{tab:p_causal_arrow} and Table \ref{tab:r_causal_arrow}, our algorithm obtain the highest precision in almost cases. Theoretically speaking, FCI cannot learn fully directed edges in these cases, so both its precision and recall are zero. Because lvLiNGAM uses overcomplete independent component analysis (OICA) to estimate the mixing matrix, it is difficult to learn the correct causal structure when the mixing matrix is estimated inaccurately. DLiNGAM is based on causal sufficient assumption, i.e., without considering the presence of latent variables, which will learn redundant and incorrect directed edges. So lvLiNGAM and DLiNGAM get the highest recall with low precision in almost all cases. The reason PLiNGAM and RCD achieve low precision and low recall is that they cannot determine the causal relationship between observed variables directly affected by latent confounders.
\begin{table}[ht]
  \caption{Precision (and its variance) of learned directed causal edges with different methods.}
  \label{tab:p_causal_arrow}
  \centering
\begin{tabular}{cccccccc}
\toprule
\multicolumn{2}{c}{Algorithm} & FCI  & lvLiNGAM & DLiNGAM & PLiNGAM & RCD  & Ours \\
\midrule
\multirow{3}{*}{Case 2} & 500  & 0.00 (0.00) & 0.23 (0.02) & 0.30 (0.01) & 0.15 (0.05) & 0.25 (0.05) & \textbf{0.60 (0.24)} \\
     & 1000 & 0.00 (0.00) & 0.10 (0.02) & 0.33 (0.00) & 0.10 (0.04) & 0.15 (0.05) & \textbf{0.70 (0.21)} \\
     & 2000 & 0.00 (0.00) & 0.20 (0.03) & 0.33 (0.00) & 0.05 (0.02) & 0.08 (0.03) & \textbf{0.50 (0.25)} \\
\midrule
\multirow{3}{*}{Case 3} & 500  & 0.00 (0.00) & 0.25 (0.01) & 0.20 (0.02) & \textbf{0.52 (0.15)} & 0.43 (0.10) & 0.47 (0.23) \\
     & 1000 & 0.00 (0.00) & 0.30 (0.00) & 0.23 (0.01) & 0.18 (0.10) & 0.23 (0.11) & \textbf{0.60 (0.24)} \\
     & 2000 & 0.00 (0.00) & 0.28 (0.01) & 0.26 (0.01) & 0.10 (0.09) & 0.20 (0.16) & \textbf{0.70 (0.21)} \\
\midrule
\multirow{3}{*}{Case 4} & 500  & 0.00 (0.00) & 0.21 (0.01) & 0.30 (0.00) & 0.09 (0.03) & 0.15 (0.04) & \textbf{0.58 (0.11)} \\
     & 1000 & 0.00 (0.00) & 0.22 (0.01) & 0.32 (0.00) & 0.05 (0.02) & 0.16 (0.04) & \textbf{0.53 (0.15)} \\
     & 2000 & 0.00 (0.00) & 0.20 (0.01) & 0.33 (0.00) & 0.00 (0.00) & 0.03 (0.01) & \textbf{0.38 (0.12)} \\
\midrule
\multirow{3}{*}{Case 6} & 500  & 0.00 (0.00) & 0.10 (0.01) & 0.17 (0.00) & 0.00 (0.00) & 0.00 (0.00) & \textbf{0.60 (0.24)} \\
     & 1000 & 0.00 (0.00) & 0.10 (0.01) & 0.10 (0.01) & 0.00 (0.00) & 0.00 (0.00) & \textbf{0.40 (0.24)} \\
     & 2000 & 0.00 (0.00) & 0.10 (0.01) & 0.12 (0.01) & 0.00 (0.00) & 0.00 (0.00) & \textbf{0.30 (0.21)} \\
\bottomrule
\end{tabular}%

\end{table}

\begin{table}[htbp]
  \caption{Recall (and its variance) of learned directed causal edges with different methods.}
  \label{tab:r_causal_arrow}
  \centering
\begin{tabular}{cccccccc}
\toprule
\multicolumn{2}{c}{Algorithm} & FCI  & lvLiNGAM & DLiNGAM & PLiNGAM & RCD  & Ours \\
\midrule
\multirow{3}{*}{Case 2} & 500  & 0.00 (0.00) & 0.70 (0.21) & \textbf{0.90 (0.09)} & 0.30 (0.21) & 0.60 (0.24) & 0.60 (0.24) \\
     & 1000 & 0.00 (0.00) & 0.30 (0.21) & \textbf{1.00 (0.00)} & 0.20 (0.16) & 0.30 (0.21) & 0.70 (0.21) \\
     & 2000 & 0.00 (0.00) & 0.60 (0.24) & \textbf{1.00 (0.00)} & 0.10 (0.09) & 0.20 (0.16) & 0.50 (0.25) \\
\midrule
\multirow{3}{*}{Case 3} & 500  & 0.00 (0.00) & \textbf{0.75 (0.11)} & 0.60 (0.14) & 0.40 (0.09) & 0.50 (0.15) & 0.30 (0.11) \\
     & 1000 & 0.00 (0.00) & \textbf{0.90 (0.04)} & 0.70 (0.06) & 0.15 (0.05) & 0.25 (0.11) & 0.30 (0.06) \\
     & 2000 & 0.00 (0.00) & \textbf{0.85 (0.05)} & 0.75 (0.11) & 0.05 (0.02) & 0.10 (0.04) & 0.35 (0.05) \\
\midrule
\multirow{3}{*}{Case 4} & 500  & 0.00 (0.00) & 0.60 (0.09) & \textbf{0.90 (0.04)} & 0.20 (0.16) & 0.35 (0.20) & 0.65 (0.15) \\
     & 1000 & 0.00 (0.00) & 0.65 (0.05) & \textbf{0.95 (0.02)} & 0.05 (0.02) & 0.30 (0.16) & 0.50 (0.15) \\
     & 2000 & 0.00 (0.00) & 0.60 (0.09) & \textbf{1.00 (0.00)} & 0.00 (0.00) & 0.05 (0.02) & 0.40 (0.14) \\
\midrule
\multirow{3}{*}{Case 6} & 500  & 0.00 (0.00) & 0.60 (0.24) & \textbf{1.00 (0.00)} & 0.00 (0.00) & 0.00 (0.00) & 0.60 (0.24) \\
     & 1000 & 0.00 (0.00) & \textbf{0.60 (0.24)} & \textbf{0.60 (0.24)} & 0.00 (0.00) & 0.00 (0.00) & 0.40 (0.24) \\
     & 2000 & 0.00 (0.00) & 0.60 (0.24) & \textbf{0.70 (0.21)} & 0.00 (0.00) & 0.00 (0.00) & 0.30 (0.21) \\
\bottomrule
\end{tabular}%

\end{table}

\begin{table}[htbp]
  \caption{Precision (and its variance) of learned non-adjacent relationships with different methods.}
  \label{tab:p_causal_nonadj}
  \centering
\begin{tabular}{cccccccc}
\toprule
\multicolumn{2}{c}{Algorithm} & FCI  & lvLiNGAM & DLiNGAM & PLiNGAM & RCD  & Ours \\
\midrule
\multirow{3}{*}{Case 1} & 500  & \textbf{1.00 (0.00)} & 0.00 (0.00) & 0.00 (0.00) & 0.60 (0.24) & 0.70 (0.21) & \textbf{1.00 (0.00)} \\
     & 1000  & 0.70 (0.21) & 0.00 (0.00) & 0.10 (0.09) & 0.30 (0.21) & 0.60 (0.24) & \textbf{1.00 (0.00)} \\
     & 2000  & 0.70 (0.21) & 0.00 (0.00) & 0.00 (0.00) & 0.00 (0.00) & 0.30 (0.21) & \textbf{1.00 (0.00)} \\
\midrule
\multirow{3}{*}{Case 2} & 500  & 0.00 (0.00) & 0.00 (0.00) & 0.00 (0.00) & 0.00 (0.00) & 0.10 (0.09) & \textbf{0.80 (0.16)} \\
     & 1000  & 0.00 (0.00) & 0.00 (0.00) & 0.00 (0.00) & 0.00 (0.00) & 0.00 (0.00) & \textbf{1.00 (0.00)} \\
     & 2000  & 0.00 (0.00) & 0.00 (0.00) & 0.00 (0.00) & 0.00 (0.00) & 0.00 (0.00) & \textbf{0.90 (0.09)} \\
\midrule
\multirow{3}{*}{Case 3} & 500  & \textbf{1.00 (0.00)} & 0.00 (0.00) & 0.10 (0.09) & 0.70 (0.21) & 0.90 (0.09) & 0.80 (0.02) \\
     & 1000  & \textbf{1.00 (0.00)} & 0.00 (0.00) & 0.00 (0.00) & 0.30 (0.21) & 0.60 (0.24) & 0.83 (0.03) \\
     & 2000  & \textbf{1.00 (0.00)} & 0.00 (0.00) & 0.20 (0.16) & 0.10 (0.09) & 0.60 (0.24) & 0.91 (0.01) \\
\midrule
\multirow{3}{*}{Case 4} & 500  & \textbf{1.00 (0.00)} & 0.10 (0.09) & 0.00 (0.00) & 0.20 (0.16) & 0.40 (0.24) & 0.96 (0.01) \\
     & 1000  & \textbf{1.00 (0.00)} & 0.00 (0.00) & 0.00 (0.00) & 0.10 (0.09) & 0.30 (0.21) & 0.96 (0.01) \\
     & 2000  & 0.80 (0.16) & 0.00 (0.00) & 0.00 (0.00) & 0.00 (0.00) & 0.00 (0.00) & \textbf{0.90 (0.09)} \\
\midrule
\multirow{3}{*}{Case 5} & 500  & \textbf{1.00 (0.00)} & 0.00 (0.00) & 0.00 (0.00) & 0.00 (0.00) & \textbf{1.00 (0.09)} & \textbf{1.00 (0.00)} \\
     & 1000  & \textbf{1.00 (0.00)} & 0.00 (0.00) & 0.00 (0.00) & 0.00 (0.00) & \textbf{1.00 (0.00)} & \textbf{1.00 (0.00)} \\
     & 2000  & \textbf{1.00 (0.00)} & 0.00 (0.00) & 0.10 (0.09) & 0.00 (0.00) & 0.90 (0.09) & \textbf{1.00 (0.00)} \\
\midrule
\multirow{3}{*}{Case 6} & 500  & 0.90 (0.09) & 0.00 (0.00) & 0.00 (0.00) & 0.00 (0.00) & 0.00 (0.00) & \textbf{0.98 (0.00)} \\
     & 1000  & 0.60 (0.24) & 0.00 (0.00) & 0.00 (0.00) & 0.00 (0.00) & 0.00 (0.00) & \textbf{1.00 (0.00)} \\
     & 2000  & 0.30 (0.21) & 0.00 (0.00) & 0.00 (0.00) & 0.00 (0.00) & 0.00 (0.00) & \textbf{1.00 (0.00)} \\
\bottomrule
\end{tabular}%

\end{table}

\begin{table}[htbp]
  \caption{Recall (and its variance) of learned non-adjacent relationships with different methods.}
  \label{tab:r_causal_nonadj}
  \centering

\begin{tabular}{cccccccc}
\toprule
\multicolumn{2}{c}{Algorithm} & FCI  & lvLiNGAM & DLiNGAM & PLiNGAM & RCD  & Ours \\
\midrule
\multirow{3}{*}{Case 1} & 500  & 0.33 (0.00) & 0.00 (0.00) & 0.00 (0.00) & 0.20 (0.03) & 0.23 (0.02) & \textbf{0.87 (0.03)} \\
     & 1000  & 0.23 (0.02) & 0.00 (0.00) & 0.03 (0.01) & 0.10 (0.02) & 0.20 (0.03) & \textbf{0.90 (0.02)} \\
     & 2000  & 0.23 (0.02) & 0.00 (0.00) & 0.00 (0.00) & 0.00 (0.00) & 0.10 (0.02) & \textbf{0.93 (0.02)} \\
\midrule
\multirow{3}{*}{Case 2} & 500  & 0.00 (0.00) & 0.00 (0.00) & 0.00 (0.00) & 0.00 (0.00) & 0.05 (0.02) & \textbf{0.80 (0.16)} \\
     & 1000  & 0.00 (0.00) & 0.00 (0.00) & 0.00 (0.00) & 0.00 (0.00) & 0.00 (0.00) & \textbf{1.00 (0.00)} \\
     & 2000  & 0.00 (0.00) & 0.00 (0.00) & 0.00 (0.00) & 0.00 (0.00) & 0.00 (0.00) & \textbf{0.90 (0.09)} \\
\midrule
\multirow{3}{*}{Case 3} & 500  & 0.60 (0.03) & 0.00 (0.00) & 0.03 (0.01) & 0.28 (0.04) & 0.33 (0.03) & \textbf{0.90 (0.02)} \\
     & 1000  & 0.53 (0.01) & 0.00 (0.00) & 0.00 (0.00) & 0.13 (0.04) & 0.23 (0.04) & \textbf{0.85 (0.03)} \\
     & 2000  & 0.45 (0.01) & 0.00 (0.00) & 0.05 (0.01) & 0.05 (0.02) & 0.23 (0.04) & \textbf{0.88 (0.02)} \\
\midrule
\multirow{3}{*}{Case 4} & 500  & 0.35 (0.02) & 0.03 (0.01) & 0.00 (0.00) & 0.05 (0.01) & 0.10 (0.02) & \textbf{0.83 (0.01)} \\
     & 1000  & 0.33 (0.01) & 0.00 (0.00) & 0.00 (0.00) & 0.03 (0.01) & 0.08 (0.01) & \textbf{0.80 (0.01)} \\
     & 2000  & 0.20 (0.01) & 0.00 (0.00) & 0.00 (0.00) & 0.00 (0.00) & 0.00 (0.00) & \textbf{0.68 (0.05)} \\
\midrule
\multirow{3}{*}{Case 5} & 500  & 0.27 (0.01) & 0.00 (0.00) & 0.00 (0.00) & 0.00 (0.00) & 0.25 (0.03) & \textbf{0.62 (0.02)} \\
     & 1000  & 0.18 (0.00) & 0.00 (0.00) & 0.00 (0.00) & 0.00 (0.00) & 0.20 (0.00) & \textbf{0.68 (0.04)} \\
     & 2000  & 0.17 (0.00) & 0.00 (0.00) & 0.02 (0.00) & 0.00 (0.00) & 0.15 (0.00) & \textbf{0.65 (0.02)} \\
\midrule
\multirow{3}{*}{Case 6} & 500  & 0.30 (0.02) & 0.00 (0.00) & 0.00 (0.00) & 0.00 (0.00) & 0.00 (0.00) & \textbf{0.92 (0.03)} \\
     & 1000  & 0.14 (0.02) & 0.00 (0.00) & 0.00 (0.00) & 0.00 (0.00) & 0.00 (0.00) & \textbf{0.96 (0.01)} \\
     & 2000  & 0.06 (0.01) & 0.00 (0.00) & 0.00 (0.00) & 0.00 (0.00) & 0.00 (0.00) & \textbf{0.96 (0.01)} \\
\bottomrule
\end{tabular}%

\end{table}

\textbf{Evaluation on non-adjacent relationships.} As shown in Table \ref{tab:p_causal_nonadj} and Table \ref{tab:r_causal_nonadj}, our algorithm obtains the highest precision and recall in almost cases. It shows that our method can accurately estimate the mixing matrix. Also with the help of the One-Latent-Component condition, the causal relationship between variables can be learned correctly. None of these methods has the ability to detect causal relationships between observed variables under the influence of latent variables, except for lvLiNGAM. But the overcomplete independent component analysis (OICA) used by lvLiNGAM usually gets stuck in local optima and further results in its low accuracy.

\begin{table}[ht]
    \centering
    \caption{Evaluation on average computation time with different methods.}
\begin{tabular}{cccccccc}
\toprule
\multicolumn{2}{c}{Algorithm} & FCI  & lvLiNGAM & DLiNGAM & PLiNGAM & RCD  & Ours \\
\midrule
\multirow{3}{*}{Case 1} & 500  & 0.10  & 3.27  & 0.08  & 0.17  & 0.18  & 0.34  \\
     & 1000 & 0.58  & 4.61  & 0.03  & 0.73  & 0.91  & 2.19  \\
     & 2000 & 4.01  & 6.63  & 0.03  & 3.53  & 5.46  & 15.51  \\
\midrule
\multirow{3}{*}{Case 2} & 500  & 0.13  & 3.38  & 0.04  & 0.12  & 0.14  & 0.53  \\
     & 1000 & 0.68  & 4.84  & 0.04  & 0.64  & 0.92  & 2.86  \\
     & 2000 & 4.18  & 6.86  & 0.02  & 3.81  & 5.70  & 18.14  \\
\midrule
\multirow{3}{*}{Case 3} & 500  & 0.43  & 64.74  & 0.04  & 0.26  & 0.38  & 1.20  \\
     & 1000 & 2.63  & 44.25  & 0.04  & 1.33  & 2.30  & 7.57  \\
     & 2000 & 16.06  & 42.90  & 0.06  & 7.18  & 14.72  & 47.27  \\
\midrule
\multirow{3}{*}{Case 4} & 500  & 0.38  & 39.04  & 0.03  & 0.23  & 0.38  & 1.07  \\
     & 1000 & 2.34  & 71.15  & 0.04  & 1.25  & 2.39  & 7.10  \\
     & 2000 & 15.80  & 67.83  & 0.08  & 7.03  & 14.80  & 49.81  \\
\midrule
\multirow{3}{*}{Case 5} & 500  & 1.25  & 207.05  & 0.04  & 0.26  & 0.38  & 3.54  \\
     & 1000 & 8.27  & 291.65  & 0.04  & 1.30  & 1.72  & 24.21  \\
     & 2000 & 45.25  & 353.70  & 0.07  & 7.07  & 10.00  & 134.99  \\
\midrule
\multirow{3}[1]{*}{Case 6} & 500  & 1.18  & 188.65  & 0.05  & 0.19  & 0.45  & 3.41  \\
     & 1000 & 7.09  & 176.42  & 0.05  & 1.13  & 2.55  & 20.85  \\
     & 2000 & 50.83  & 334.49  & 0.05  & 6.93  & 15.12  & 155.45  \\
\bottomrule
\end{tabular}%
    \label{tab:time}
\end{table}
\begin{table}[htbp]
    \centering
    \caption{Evaluation on average RMSE from synthetic data.}

\begin{tabular}{ccccc}
\toprule
\multicolumn{2}{c}{Algorithm} & lvLiNGAM & DLiNGAM & Ours \\
\midrule
\multirow{3}{*}{Case 1} & 500  & 6.38 (328.03) & 0.23 (0.00) & 0.05 (0.00) \\
     & 1000 & 0.53 (0.14) & 0.23 (0.00) & 0.00 (0.00) \\
     & 2000 & 2.44 (37.07) & 0.23 (0.01) & 0.00 (0.00) \\
\midrule
\multirow{3}{*}{Case 2} & 500  & 0.34 (0.03) & 0.28 (0.02) & 0.04 (0.00) \\
     & 1000 & 2.86 (16.87) & 0.27 (0.01) & 0.03 (0.00) \\
     & 2000 & 0.83 (0.46) & 0.27 (0.02) & 0.04 (0.00) \\
\midrule
\multirow{3}{*}{Case 3} & 500  & 1.66 (3.36) & 0.28 (0.02) & 0.16 (0.01) \\
     & 1000 & 0.70 (0.11) & 0.23 (0.00) & 0.24 (0.09) \\
     & 2000 & 1.83 (7.13) & 0.23 (0.01) & 0.12 (0.00) \\
\midrule
\multirow{3}{*}{Case 4} & 500  & 19.34 (2792.92) & 0.30 (0.03) & 0.07 (0.00) \\
     & 1000 & 5.59 (98.18) & 0.28 (0.02) & 0.08 (0.00) \\
     & 2000 & 0.57 (0.12) & 0.25 (0.01) & 0.08 (0.00) \\
\midrule
\multirow{3}{*}{Case 5} & 500  & 1.27 (3.29) & 0.19 (0.01) & 0.10 (0.01) \\
     & 1000 & 1.39 (2.77) & 0.26 (0.00) & 0.21 (0.04) \\
     & 2000 & 7.28 (200.35) & 0.26 (0.01) & 0.09 (0.01) \\
\midrule
\multirow{3}{*}{Case 6} & 500  & 9.30 (269.89) & 0.25 (0.01) & 0.02 (0.00) \\
     & 1000 & 4.96 (117.11) & 0.27 (0.01) & 0.03 (0.00) \\
     & 2000 & 1.15 (0.65) & 0.28 (0.01) & 0.03 (0.00) \\
\bottomrule
\end{tabular}%

    \label{tab:rmse}
\end{table}

\textbf{Evaluation on average computation time.} Table \ref{tab:time} illustrates the average computation time of each method. Compared with lvLiNGAM, the average computation time is lower than that of lvLiNGAM, especially when the sample size is 500. Notice that the cases shown in Figure \ref{fig:exp_graphs} contain at most four observed variables. If the number of observed variables increases, the computation time of lvLiNGAM will be very high. Thus, these experiment results on average computation time validate the effectiveness of our methods compared with lvLiNGAM. 

\textbf{Evaluation on the causal coefficient estimation.} 
Because FCI only returns the causal structure without causal coefficient, and the results of PLiNGAM and RCD contain many or all ``Nan" values, for fairness, we only provide the RMSE of lvLiNGAM, DLiNGAM, and our methods which are illustrated in Table \ref{tab:rmse}. From the results, we can see that the RMSE of our method is the smallest, and the variance of the RMSE is close to zero. This reflects the reliability and robustness of our method on causal coefficient estimation by higher-order cumulants. The RMSE and its variance of lvLiNGAM vary greatly because the lvLiNGAM easily gets stuck in local optima. The variances of RMSE of DLiNGAM are also close to zero, but the RMSE of DLiNGAM is larger than ours in most cases. Since DLiNGAM returns the dense graphs, these graphs contain the causal coefficient estimation of the true causal edges.

\subsection{Synthetic Data in a More Easy-to-learn Setting}\label{app:exp_syn}

Besides, we conducted more experiments to show the performance of our algorithm in a more simple, easy-to-learn setting. In detail, we designed the Case 7 over six observed variables $\lbrace X_1, \dots, X_6 \rbrace$ and two latent variables $\lbrace L_1, L_2 \rbrace$ as follows:

[Case 7]: latent variable $L_1$ has six observed variables as children, i.e., $L_1 \rightarrow \lbrace X_1, X_2, X_3, X_4, X_5, X_6 \rbrace$; the other latent variable $L_2$ directly causes three observed variables $X_4$, $X_5$ and $X_6$, i.e., $L_2 \rightarrow \lbrace X_4, X_5, X_6 \rbrace$; observed variable $X_2$ directly causes two observed variables $X_3$ and $X_4$, i.e., $X_2 \rightarrow \lbrace X_3, X_4 \rbrace$; observed variable $X_3$ directly causes two observed variables $X_4$ and $X_5$, i.e., $X_3 \rightarrow \lbrace X_4, X_5 \rbrace$; observed variable $X_4$ directly causes observed variable $X_5$, i.e., $X_4 \rightarrow X_5$.

\begin{table}[htbp]
  \centering
  \caption{Metrics of learned directed causal edges with different methods in the Case  7.}
    \begin{tabular}{cccccccc}
    \toprule
          & Sample Size & FCI   & lvLiNGAM & DLiNGAM & PLiNGAM & RCD   & Ours \\
    \midrule
    \multirow{3}[2]{*}{Precision} & 500   & 0.00  & -     & 0.07  & 0.16  & 0.33  & 0.86  \\
          & 1000  & 0.00  & -     & 0.10  & 0.11  & 0.00  & 0.85  \\
          & 2000  & 0.00  & -     & 0.15  & 0.00  & 0.06  & 0.87  \\
    \midrule
    \multirow{3}[2]{*}{Recall} & 500   & 0.00  & -     & 0.22  & 0.14  & 0.10  & 0.42  \\
          & 1000  & 0.00  & -     & 0.30  & 0.10  & 0.00  & 0.50  \\
          & 2000  & 0.00  & -     & 0.46  & 0.00  & 0.02  & 0.64  \\
    \midrule
    \multirow{3}[2]{*}{F1-score} & 500   & 0.00  & -     & 0.11  & 0.14  & 0.15  & 0.55  \\
          & 1000  & 0.00  & -     & 0.15  & 0.10  & 0.00  & 0.61  \\
          & 2000  & 0.00  & -     & 0.23  & 0.00  & 0.03  & 0.73  \\
    \bottomrule
    \end{tabular}%
  \label{tab:easy_arrow}%
\end{table}%

\begin{table}[htbp]
  \centering
  \caption{Metrics of learned non-adjacent relationships with different methods in the Case  7.}
    \begin{tabular}{cccccccc}
    \toprule
          & Sample Size & FCI   & lvLiNGAM & DLiNGAM & PLiNGAM & RCD   & Ours \\
    \midrule
    \multirow{3}[2]{*}{Precision} & 500   & 0.00  & -     & 0.00  & 0.17  & 0.41  & 0.86  \\
          & 1000  & 0.00  & -     & 0.00  & 0.17  & 0.00  & 0.90  \\
          & 2000  & 0.00  & -     & 0.10  & 0.00  & 0.05  & 0.95  \\
    \midrule
    \multirow{3}[2]{*}{Recall} & 500   & 0.00  & -     & 0.00  & 0.04  & 0.13  & 0.92  \\
          & 1000  & 0.00  & -     & 0.00  & 0.03  & 0.00  & 0.92  \\
          & 2000  & 0.00  & -     & 0.01  & 0.00  & 0.01  & 0.93  \\
    \midrule
    \multirow{3}[2]{*}{F1-score} & 500   & 0.00  & -     & 0.00  & 0.06  & 0.19  & 0.88  \\
          & 1000  & 0.00  & -     & 0.00  & 0.05  & 0.00  & 0.91  \\
          & 2000  & 0.00  & -     & 0.02  & 0.00  & 0.02  & 0.94  \\
    \bottomrule
    \end{tabular}%
  \label{tab:easy_non_adj}%
\end{table}%

Unfortunately, the lvLiNGAM cannot return the results of one generated dataset in 5 days. So we only provide the results of our method and compared methods except for lvLiNGAM in Table \ref{tab:easy_arrow} - \ref{tab:easy_non_adj}. From the result, we observed that if the number of directed edges is larger than that of the cases in the original manuscript, the F1-score of the learned directed causal edges by our method is much higher than that in the original manuscript. But the F1-scores of other methods are still very low because they cannot recover the structure that all observed variables are directly affected by the same latent variables.

\subsection{Synthetic Data in Three Settings of Assumptions Violation}\label{app:exp_asm}

In order to evaluate the behavior of the procedure when various assumptions are violated, we further conducted experiments on synthesis data. In detail, we designed the following three cases over six observed variables $\{X_1, X_2, \dots, X_6\}$ and two latent variables $\{L_1, L_2\}$ based on Case 7. 

[Case 8]: We remove the edge $L_2 \rightarrow X_4$ from the graph in Case 7. 

[Case 9]: We further remove the edge $X_3 \rightarrow X_5$ from the graph in Case 7, and then add the edge $X_5 \rightarrow X_6$ in the graph. 

[Case 10]: We further remove the edge $X_2 \rightarrow X_4$ from the graph in Case 7, and then add the edge $X_1 \rightarrow X_2$ in the graph.

Accordingly, the properties of different cases are as follows:

1. Case 7 satisfies all the assumptions;

2. The local structure over one latent variable in Case 8 violates assumption 1;

3. The local structure over one latent variable in Case 9 violates assumption 2;

4. All local structure over two latent variables in Case 10 violates all assumptions.

The F1-scores of the learned directed causal edges and non-adjacent relationships by our method with different cases are given in Table \ref{tab:asm_violation_directed} - \ref{tab:asm_violation_non_adj}. From the result, we observed the output of the proposed procedure may contain undirected links, which implies that some of the assumptions are violated and we do not know the relationships between observed variables. That is, the algorithm can identify the causal structure over subsets of the variables for which the assumptions are satisfied (under random errors because of finite samples).

\begin{table}[t]
  \centering
  \caption{Metrics of learned directed causal edges in the cases violate the assumptions.}
    \begin{tabular}{cccccc}
    \toprule
           & Sample Size & Case 7 & Case 8 & Case 9 & Case 10 \\
    \midrule
    \multirow{3}[2]{*}{Precision} & 500   & 0.86  & 0.67  & 0.43  & 0.10  \\
          & 1000  & 0.85  & 0.58  & 0.30  & 0.00  \\
          & 2000  & 0.87  & 0.63  & 0.23  & 0.00  \\
    \midrule
    \multirow{3}[2]{*}{Recall} & 500   & 0.42  & 0.34  & 0.20  & 0.02  \\
          & 1000  & 0.50  & 0.30  & 0.18  & 0.00  \\
          & 2000  & 0.64  & 0.32  & 0.16  & 0.00  \\
    \midrule
    \multirow{3}[2]{*}{F1-score} & 500   & 0.55  & 0.43  & 0.25  & 0.03  \\
          & 1000  & 0.61  & 0.39  & 0.21  & 0.00  \\
          & 2000  & 0.73  & 0.41  & 0.19  & 0.00  \\
    \bottomrule
    \end{tabular}%
  \label{tab:asm_violation_directed}%
\end{table}%

\begin{table}[t]
  \centering
  \caption{Metrics of learned non-adjacent relationships in the cases violate the assumptions.}
    \begin{tabular}{cccccc}
    \toprule
          &  Sample Size & Case 7 & Case 8 & Case 9 & Case 10 \\
    \midrule
    \multirow{3}[2]{*}{Precision} & 500   & 0.86  & 0.90  & 0.82  & 0.35  \\
          & 1000  & 0.90  & 0.96  & 0.83  & 0.00  \\
          & 2000  & 0.95  & 0.87  & 0.85  & 0.00  \\
    \midrule
    \multirow{3}[2]{*}{Recall} & 500   & 0.92  & 0.90  & 0.65  & 0.07  \\
          & 1000  & 0.92  & 0.86  & 0.57  & 0.00  \\
          & 2000  & 0.93  & 0.93  & 0.60  & 0.00  \\
    \midrule
    \multirow{3}[2]{*}{F1-score} & 500   & 0.88  & 0.90  & 0.71  & 0.12  \\
          & 1000  & 0.91  & 0.90  & 0.67  & 0.00  \\
          & 2000  & 0.94  & 0.89  & 0.69  & 0.00  \\
    \bottomrule
    \end{tabular}%
  \label{tab:asm_violation_non_adj}%
\end{table}%

\begin{table}[htbp]
    \centering
    \caption{Metrics (and its variance) of learned causal directed edges with randomly generated graphs. }
\begin{tabular}{cccccccc}
\toprule
     & Sample Size & FCI  & lvLiNGAM & DLiNGAM & PLiNGAM & RCD  & Ours \\
\midrule
\multirow{3}[2]{*}{Precision} & 500  & 0.00 (0.00) & -    & 0.17 (0.00) & 0.00 (0.00) & 0.35 (0.20) & \textbf{0.60 (0.24)} \\
     & 1000 & 0.00 (0.00) & -    & 0.19 (0.00) & 0.00 (0.00) & 0.10 (0.09) & \textbf{0.62 (0.19)} \\
     & 2000 & 0.00 (0.00) & -    & 0.18 (0.00) & 0.00 (0.00) & 0.00 (0.00) & \textbf{0.77 (0.16)} \\
\midrule
\multirow{3}[2]{*}{Recall} & 500  & 0.00 (0.00) & -    & \textbf{0.87 (0.03)} & 0.00 (0.00) & 0.17 (0.05) & 0.23 (0.05) \\
     & 1000 & 0.00 (0.00) & -    & \textbf{0.93 (0.02)} & 0.00 (0.00) & 0.03 (0.01) & 0.37 (0.08) \\
     & 2000 & 0.00 (0.00) & -    & \textbf{0.90 (0.02)} & 0.00 (0.00) & 0.00 (0.00) & 0.43 (0.07) \\
\midrule
\multirow{3}[2]{*}{F1-score} & 500  & 0.00 (0.00) & -    & 0.29 (0.00) & 0.00 (0.00) & 0.22 (0.08) & \textbf{0.33 (0.08)} \\
     & 1000 & 0.00 (0.00) & -    & 0.31 (0.00) & 0.00 (0.00) & 0.05 (0.02) & \textbf{0.45 (0.10)} \\
     & 2000 & 0.00 (0.00) & -    & 0.30 (0.00) & 0.00 (0.00) & 0.00 (0.00) & \textbf{0.54 (0.09)} \\
\bottomrule
\end{tabular}

    \label{tab:arrowadd}
\end{table}

\begin{table}[htbp]
    \centering
    \caption{Metrics of learned causal non-adjacent relationships with randomly generated graphs. }
\begin{tabular}{cccccccc}
\toprule
     & Sample Size & FCI  & lvLiNGAM & DLiNGAM & PLiNGAM & RCD  & Ours \\
\midrule
\multirow{3}[2]{*}{Precision} & 500  & \textbf{1.00 (0.00)} & -    & 0.00 (0.00) & 0.00 (0.00) & \textbf{1.00 (0.00)} & 0.88 (0.00) \\
     & 1000 & \textbf{1.00 (0.00)} & -    & 0.10 (0.09) & 0.00 (0.00) & \textbf{1.00 (0.00)} & 0.90 (0.01) \\
     & 2000 & \textbf{1.00 (0.00)} & -    & 0.10 (0.09) & 0.00 (0.00) & \textbf{1.00 (0.00)} & 0.91 (0.00) \\
\midrule
\multirow{3}[2]{*}{Recall} & 500  & 0.49 (0.01) & -    & 0.00 (0.00) & 0.00 (0.00) & 0.83 (0.04) & \textbf{0.92 (0.01)} \\
     & 1000 & 0.31 (0.00) & -    & 0.01 (0.00) & 0.00 (0.00) & 0.70 (0.02) & \textbf{0.89 (0.00)} \\
     & 2000 & 0.29 (0.00) & -    & 0.01 (0.00) & 0.00 (0.00) & 0.57 (0.02) & \textbf{0.85 (0.01)} \\
\midrule
\multirow{3}[2]{*}{F1-score} & 500  & 0.65 (0.01) & -    & 0.00 (0.00) & 0.00 (0.00) & 0.89 (0.02) & \textbf{0.90 (0.00)} \\
     & 1000 & 0.47 (0.00) & -    & 0.02 (0.00) & 0.00 (0.00) & 0.81 (0.01) & \textbf{0.89 (0.00)} \\
     & 2000 & 0.45 (0.00) & -    & 0.02 (0.00) & 0.00 (0.00) & 0.71 (0.01) & \textbf{0.87 (0.00)} \\
\bottomrule
\end{tabular}

    \label{tab:nonadjadd}
\end{table}

\begin{table}[htbp]
    \centering
    \caption{The stocks corresponding to the observed variables.}

\begin{tabular}{cc}
\toprule
Observed Variable & Stock\\
\midrule
$X_1$ & Cheung Kong (0001.hk) \\
$X_2$ & CLP Hldgs (0002.hk) \\
$X_3$ & HK China Gas (0003.hk) \\
$X_4$ & Wharf (Hldgs) (0004.hk) \\
$X_5$ & HSBC Hldg (0005.hk) \\
$X_6$ & HK Electric (0006.hk) \\
$X_7$ & Hang Lung Dev (0010.hk) \\
$X_8$ & Hang Seng Bank (0011.hk) \\
$X_9$ & Henderson Land (0012.hk) \\
$X_{10}$ & Hutchison (0013.hk) \\
$X_{11}$ & Sun Hung Kai Prop (0016.hk) \\
$X_{12}$ & Swire Pacific ’A’ (0019.hk) \\
$X_{13}$ & Bank of East Asia (0023.hk) \\
$X_{14}$ & Cathay Pacific Air (0293.hk) \\
\bottomrule
\end{tabular}%

    \label{tab:stock}
\end{table}

\subsection{Synthetic Data Generated by Random Graphs}\label{app:exp_random}
We conducted more experiments to show the performance of our algorithm for randomly generated graphs and more variables. In detail, we randomly generated the causal graphs with 6 observed variables according to the latent variables model. Unfortunately, the lvLiNGAM cannot return the results of one generated dataset in 5 days. So we only provide the results of our method and compared methods except lvLiNGAM in Table \ref{tab:arrowadd}-\ref{tab:nonadjadd}.

\subsection{Stock Market Data}\label{app:stock}
We applied our method to the Hong Kong Stock Market Data, aiming at discovering the causal relationships among the selected 14 stocks. The data contains 1331 daily returns. 

The observed stock variables are given in Table \ref{tab:stock}.

We use the expert knowledge of the stock market and the results from two papers \citep{zhang2008minimal, cai2019triad} as references to evaluate the output of our proposed algorithm, and we find the following discoveries are consistent with the expert knowledge. In detail, the found causal relationships are: 

1) all observed variables are affected by a latent confounder. Besides, all observed variables except $X_7$ and $X_{14}$ are influenced by another latent confounder at the same time; 

2) the directed causal edges between observed variables are: $X_2 \to \{X_3, X_8 \}$, $X_9 \to X_4$, $X_{11} \to X_{10} \to X_1$; 3) the undirected edges are $X_1 - X_4$, $X_2 -X_6$, $X_1 -  X_{13} - X_{14}$. 

This estimated causal structure inspires us the findings as follows: 

1) the whole market environment is affected by the hidden variable (which may be government policy, the total risk in the market and so on) \citep{cai2019triad}. This is also consistent with the expert knowledge of the stock market. 

2) There is often a connection between stocks belonging to the same sub-index. For example, $X_2$, $X_3$ and $X_6$ are dependent because they are constituent stocks under the Hang Seng Utilities Index. 

3) The edge $X_1 \to X_{10} $ is consistent with the knowledge that ownership relations ($X_1$ holds about 50\% of $X_{10}$) tend to cause causal relations \citep{zhang2008minimal}. 

4) If the principal activity of company A is a subset of that of company B, there usually exists a causal edge from A to B. For example, the result includes $X_9 \rightarrow X_4$; in fact, real estate is the principal activity for $X_9$, while $X_4$ has principal activities in real estate, transportation, and retail. Similarly for the relationships between $X_{10}$ and $X_{11}$. 

5) If the principal activities of two companies partially overlap, then we usually end up with an undirected link between them. For example, this is the case for $X_1$ and $X_4$; in fact, the activities of $X_1$ include residential, office, retail, industrial, and hotel properties, and the activities of $X_4$ include port and related services, real estate and hotels, and department stores and retail. 

6) Banks tend to be effects of other related companies; we suspect those companies are the bank's clients and hence their returns are also reflected in the bank's return in some way (we treat this only as a hypothesis).  For example, the result includes $X_2 \rightarrow X_8$; $X_8$ is one of the largest banks in Hong Kong, and $X_2$ may be a client of $X_8$.

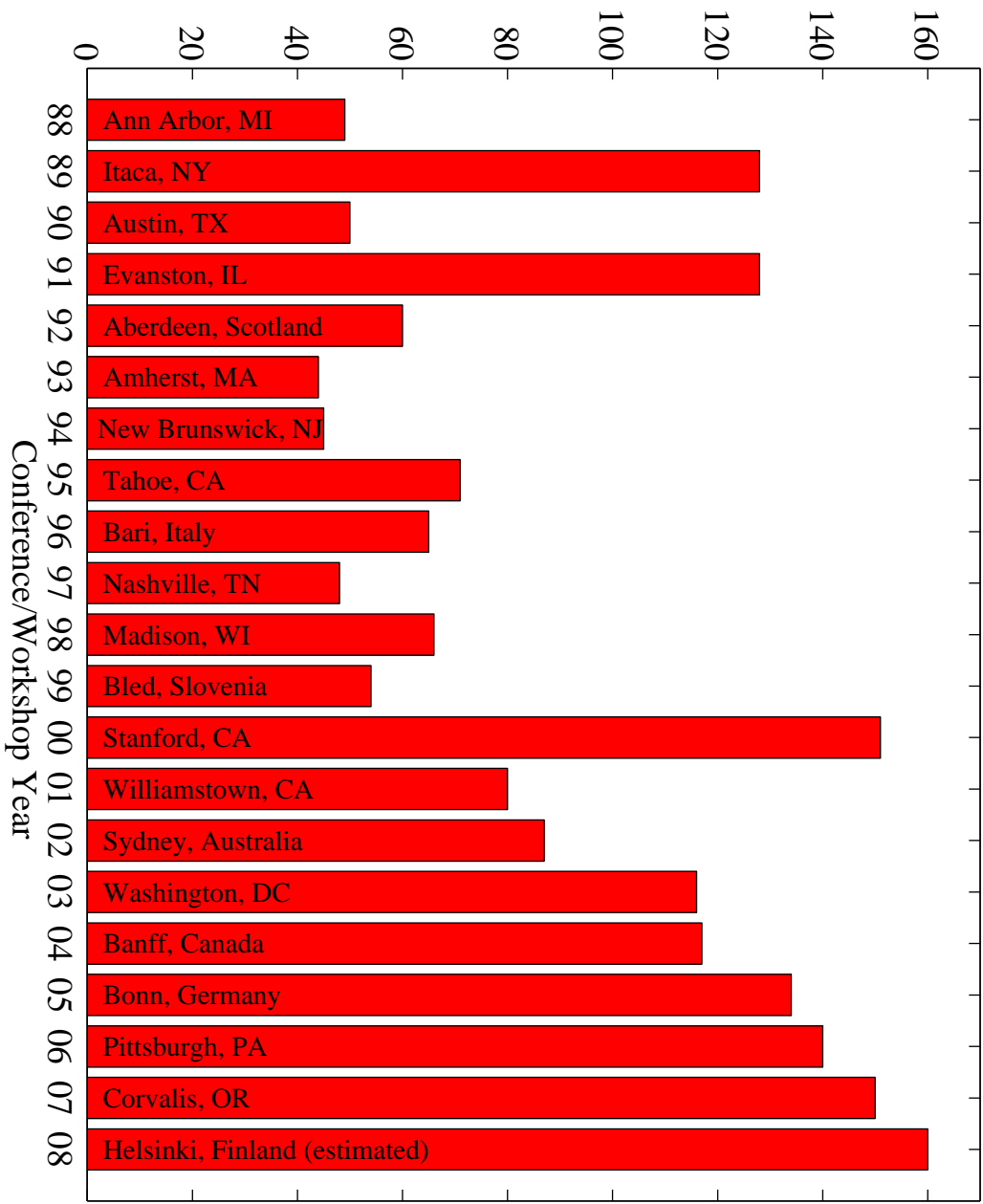
\end{document}